\documentclass[11pt]{article}
\usepackage[utf8]{inputenc}
\usepackage[T1]{fontenc}
\usepackage{lmodern}
\usepackage{amsmath}
\usepackage{amssymb}
\usepackage{graphicx}
\usepackage{booktabs}
\usepackage{array}
\usepackage{longtable}
\usepackage{multirow}
\usepackage[margin=1in]{geometry}
\usepackage{caption}
\usepackage[hidelinks]{hyperref}
\usepackage{xurl}
\usepackage{xcolor}
\usepackage{enumitem}
\usepackage{float}

\title{Inclusive Federated Learning Through Compliance-Weighted Noise Allocation in Healthcare AI}

\author{
\begin{minipage}{0.9\textwidth}\centering
Santhosh Parampottupadam\textsuperscript{1}, Melih Co\c{s}\u{g}un\textsuperscript{3}, Sarthak Pati\textsuperscript{5}, Maximilian Zenk\textsuperscript{1,2}, Saikat Roy\textsuperscript{1,2}, Dimitrios Bounias\textsuperscript{1,2}, Benjamin Hamm\textsuperscript{1,2}, Sinem Sav\textsuperscript{3}, Klaus Maier-Hein\textsuperscript{1,2,4}, Ralf Floca\textsuperscript{1}
\end{minipage}\\[1.2em]
\normalsize
\begin{minipage}{0.86\textwidth}
\centering
\textsuperscript{1}\,German Cancer Research Center (DKFZ), Division of Medical Image Computing, Heidelberg, Germany\\
\textsuperscript{2}\,Medical Faculty Heidelberg, Heidelberg University, Heidelberg, Germany\\
\textsuperscript{3}\,Department of Computer Engineering, Bilkent University, Universiteler 06800 \c{C}ankaya/Ankara, T\"urkiye\\
\textsuperscript{4}\,Pattern Analysis and Learning Group, Department of Radiation Oncology, Heidelberg University Hospital, 69120 Heidelberg, Germany\\
\textsuperscript{5}\,Medical Research Group, MLCommons, San Francisco, CA, USA
\end{minipage}
}

\date{}

\begin{document}
\maketitle
\renewcommand{\thefootnote}{}\footnotetext{Preprint. Under review at JMIR AI. Code: \href{https://github.com/santhoshcameo/inclusive-privacy-with-compliance-fl}{github.com/santhoshcameo/\allowbreak inclusive-privacy-\allowbreak with-compliance-fl}}\renewcommand{\thefootnote}{\arabic{footnote}}

\section*{Abstract}
\noindent\textbf{Background:} Federated learning (FL) enables collaborative training of clinical AI models without centralizing patient data, but adoption in healthcare is limited by privacy concerns, heterogeneous institutional compliance, and resource disparities. Standard differential privacy (DP) applies uniform noise to all clients, which can disproportionately penalize well-compliant or under-resourced institutions and reduce model performance.

\noindent\textbf{Objective:} This study introduces a compliance-aware FL framework that adapts DP mechanisms based on institutional compliance. Our goals were to (1) design a compliance scoring tool aligned with healthcare and security standards, (2) integrate adaptive DP noise into FL, and (3) evaluate its effectiveness in balancing utility and compliance under a bounded server-side privacy budget.

\noindent\textbf{Methods:} We developed a compliance scoring tool aligned with HIPAA, GDPR, NIST, ISO, and HL7/FHIR standards, mapping each client's score to a per-step Gaussian noise scale $\sigma_i$ for server-side DP-SGD on a small aggregator dataset. The mechanism provides a formal $(\varepsilon,\delta)$ bound on the aggregator dataset under a semi-honest aggregator threat model; client-level formal DP would require composition with secure aggregation, which we identify as a concrete future research direction. We evaluated five FL strategies (FedAvg, FedMedian, FedYogi, FedProx, FedAdam) on PneumoniaMNIST (n=5,856) and BreastMNIST (n=780), with 16 clients, 50 rounds, and five random seeds per configuration for per-strategy paired analysis. The cumulative aggregator-dataset $\varepsilon$ under full sequential R\'enyi-DP composition (Opacus accountant) is $\approx$1434 (BreastMNIST) and $\approx$513 (PneumoniaMNIST) at $\delta = 10^{-5}$ (per-client-epoch cost $\varepsilon \in [2, 6]$).

\noindent\textbf{Results:} Including the 12 lower-compliance clients (Experiment 1) versus the compliant-only baseline (Experiment 4) changed BreastMNIST accuracy by +4.5 (FedAvg), +6.8 (FedMedian), +5.2 (FedProx), +1.6 (FedYogi), and $-4.1$ (FedAdam) percentage points (pooled mean +2.8 pp; no per-strategy comparison reached significance at n=5 seeds; individual configurations gained up to +17 pp). Compliance-weighted allocation matched uniform server-side DP at equal mean noise scale (pooled difference +0.1 pp), indicating that the governance flexibility of compliance-based noise assignment carries no utility penalty. Introducing $\sigma_{\min}$ noise into first-round client updates cost on average 1.3 pp (BreastMNIST) and 2.5 pp (PneumoniaMNIST, FedAvg).

\noindent\textbf{Conclusions:} Compliance-weighted server-side DP enables lower-compliance institutions to participate in federated training without degrading model performance, while giving investigators an auditable, compliance-justified control over per-site noise allocation at no utility cost; formal $(\varepsilon,\delta)$ guarantees apply to the aggregator dataset, and client-level formal DP requires composition with secure aggregation. Further validation on full-resolution clinical datasets, larger seed counts with per-strategy paired tests, and inter-rater validation of the compliance scoring tool are concrete next steps before clinical deployment.

\noindent\textbf{Trial Registration:} Not applicable.

\noindent\textbf{Keywords:} federated learning; differential privacy; compliance; healthcare AI; privacy-preserving machine learning; inclusive AI

\section*{Introduction}
Artificial Intelligence (AI) can advance healthcare through improved diagnostics and personalized treatments, but privacy concerns and regulatory constraints limit its adoption. Federated learning (FL) [1] enables decentralized model training, keeping patient data within each institution while supporting collaborative clinical AI development. Despite its potential, FL in healthcare [2--4] faces challenges in data security, privacy, and inclusivity. FL systems are vulnerable to reconstruction attacks, where model updates can reveal sensitive information [5,6]. Differential privacy (DP), which adds calibrated noise to reduce the risk that individual records can be inferred from model updates, has been integrated into FL to mitigate these risks, providing theoretical guarantees against data reconstruction and inference attacks [7,8]. However, DP introduces trade-offs by adding noise to model updates, often degrading performance [9,10,39]. Traditional DP methods apply equal amounts of noise for each client [11] (which we refer to as Uniform DP), overlooking disparities such as compliance and resources [12,13].

In this manuscript, the term ``privacy'' refers specifically to data privacy in collaborative model training---i.e., the protection of clients' local training data against (i) gradient-inversion and reconstruction attacks on transmitted updates [5,6], (ii) memorisation and post-training extraction attacks on the released global model [48], and (iii) statistical inference attacks (including membership inference) by an honest-but-curious aggregator. The broader notions of clinical privacy, patient-consent management, regulated disclosure to care providers, longitudinal de-identification are operationally relevant inputs to our compliance scoring tool but are not themselves the protection target of the DP mechanism we propose. The compliance-weighted adaptive DP mechanism in this paper adapts the Gaussian noise scale at the aggregator to each client's institutional readiness; it bounds the leakage of the released global model with respect to the aggregator dataset, and protection of raw client updates in transit is a transport-layer property requiring secure aggregation (see Threat Model).

Healthcare FL faces significant challenges due to institutional heterogeneity [40, 41], with DP imposing high computational demands that often require specialized hardware [14--17]. Clinical sites with lower patient loads struggle to participate due to resource constraints, compliance gaps, and coordination overhead [16,18,19]. Real-world FL studies [20,21] demonstrate feasibility but rely on trust-based federations, marginalizing smaller institutions. Balancing privacy and utility in DP requires clear trade-offs, as any DP implementation impacts model performance. A review of 612 studies [19] found that 32 of 612 (5.2\%) involved real-world clinical applications, highlighting the need for FL frameworks that ensure privacy, inclusivity, and equitable participation while addressing compliance and computational barriers [17,19,22]. This framework is demonstrated in this work on medical-imaging data (chest-radiograph and breast-ultrasound classification benchmarks). The mechanism itself --- compliance scoring, the Eq. 1/Eq. 2 mappings, and server-side DP-SGD --- is modality-agnostic; generalisation to electronic health records, genomic data, time-series, and multimodal medical data is identified as future work.

Inclusion in federated learning is not merely a quantitative concern. Across countries, continents, and resource tiers, the institutions that would benefit most from cross-site collaboration, regional hospitals, public-health programmes in low- and middle-income settings, emerging-market clinics, and specialist centres treating under-represented patient populations---are often the same institutions whose compliance infrastructure lags behind that of mature academic medical centres. A uniform-DP regime forces federation organisers into a binary choice for each candidate site: include the site under the same noise scale as everyone else (potentially insufficient for higher-risk sites), or exclude the site (sacrificing the diversity of its patient population). Either choice replicates the structural exclusion documented in clinical FL reviews [19] and produces models that under-represent the populations most likely to encounter them. The framework presented here dissolves this binary by allocating noise proportional to institutional compliance: lower-compliance sites are assigned larger noise scales in the server-side DP step, higher-compliance sites incur less noise, and federation participation expands without relaxing the noise scales assigned to the most-exposed participants.

To address these challenges, our approach rethinks how differential privacy is applied in federated learning. As illustrated in Figure 1, existing FL implementations that apply uniform DP either at the client side (Figure 1a) or post-aggregation (Figure 1b) often exclude low-compliance or resource-limited institutions that cannot meet the hardware or regulatory prerequisites for DP. In contrast, our framework (Figure 1c) enables these otherwise-excluded clients to participate under an adaptive, compliance-aware DP mechanism. Instead of applying uniform noise---which can degrade performance for well-compliant institutions and reduce the utility of the global model---we allocate noise proportional to each client's calculated compliance score, which we define as a normalized value between 0 and 1 summarizing selected institutional readiness factors. This ensures that even low-compliance participants contribute under a consistent noise-allocation policy, while high-compliance clients are not over-penalized. Ultimately, our goal is to enhance inclusivity and model performance, bridging the gap between privacy protection and practical deployment in heterogeneous clinical environments (evaluated in this study on low-resolution benchmark data).

The framework introduces a customizable compliance scoring tool aligned with established healthcare privacy, security, and interoperability standards while maintaining inclusivity. It incorporates privacy concepts from various regulatory and best-practice frameworks such as patient consent management [23], anonymization practices [24,25], audit logs \& network security [26], data encryption \& secure infrastructure [27], ethical AI policies [28], interoperability [29], and data \& model training quality. These standards collectively address privacy risks, enforce secure data handling, and promote equitable FL scalability in clinical environments.

Under the semi-honest aggregator threat model described in the Methods (see Threat Model subsection), our framework performs adaptive server-side DP, optimizing noise injection to balance privacy and utility [30]. By adapting noise levels to client compliance scores, it maintained performance in our benchmark settings. The compliance scoring tool enables investigators to weigh regulatory adherence, data integrity, and security protocols, fostering tailored and trustworthy FL deployments. We evaluated our method on multiple public datasets [31] and aggregation methods [1,32,33], and quantified accuracy changes relative to compliant-only baselines at matched mean noise scale (individual configurations gained up to +17 percentage points in the released reference implementation).

This manuscript's contributions are:
\begin{itemize}[leftmargin=*]
\item A compliance-aware FL framework with adaptive DP, adjusting noise based on client compliance to enhance fairness and inclusivity.
\item A web-based compliance scoring tool --- to our knowledge, the first research tool of its kind for federated-learning compliance quantification --- aligned with HIPAA, GDPR, NIST, ISO, and HL7/FHIR standards. The tool is fully PI-configurable (factor selection, weight assignment, and the eligibility-gate-versus-modulating partition are all user-editable) and will be released as open source upon acceptance, enabling independent inter-rater validation studies and community-driven refinement.
\item Implementation of adaptive server-side DP, enabling resource-constrained clinics to participate while balancing privacy and performance.
\end{itemize}

\begin{figure}[H]
\centering
\includegraphics[width=\linewidth]{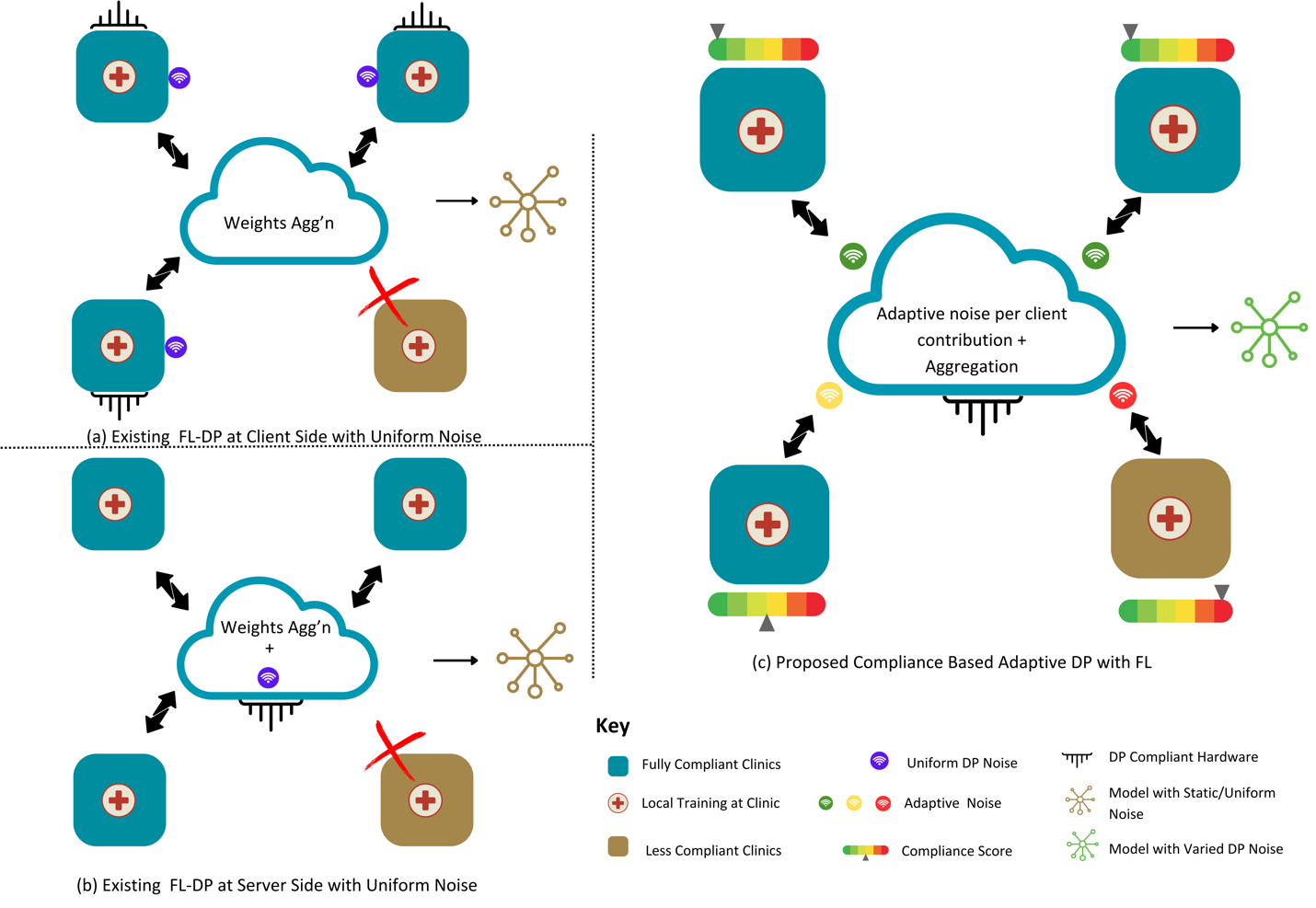}
\caption*{Figure 1. (a) Existing FL with client-side DP uses uniform noise, requiring DP-compliant hardware, limiting less compliant, resource-constrained clinics. (b) Server-side DP adds uniform noise post-aggregation, reducing privacy-utility efficiency and further excluding less compliant clinics. (c) Our compliance-aware adaptive DP applies per-client noise before aggregation, enabling participation from low-resource, less compliant clinics, allocating server-side DP noise by compliance score while preserving performance.}
\end{figure}

\section*{Methods}
The remainder of this section is organised to address each of the three stated objectives in turn --- the compliance scoring tool, adaptive DP integration, and the evaluation --- and concludes with Ethical Considerations.

\subsection*{Notation}
The following symbols are used throughout Equations 1--2 and Algorithm 1. $n$ = number of compliance factors selected by the PI for site $i$. $w_i$ = weight of compliance factor $i$ (weights are renormalised so the per-site weighted average is well-defined). $s_i$ = option score for factor $i$, $s_i \in [0, 1]$. $S_c$ = compliance score of a client, $S_c = \sum_i w_i s_i / \sum_i w_i$, with $S_c \in [0, 1]$. $\sigma_i$ = Gaussian noise scale applied to client $i$'s contribution at the aggregator. $\sigma_{\min}$ = minimum noise scale ($\sigma_{\min} = 0.4$ in this study). $\sigma_{\max}$ = maximum noise scale ($\sigma_{\max} = 1.0$). $\varepsilon$ = cumulative privacy budget (estimated via Opacus / R\'enyi-DP composition). $\delta$ = DP failure probability ($\delta = 10^{-5}$; satisfies $\delta < 1/n_\text{train}$ for both PneumoniaMNIST and BreastMNIST).

\subsection*{Threat Model}
The mechanism operates under a semi-honest (honest-but-curious) aggregator assumption: the central server is trusted to execute the protocol correctly --- applying the per-client DP-SGD step on the aggregator dataset and forwarding the resulting noisy updates to the chosen aggregation rule --- but may attempt to learn about client data from information it legitimately observes (raw post-training client weights, intermediate updates, and the final aggregated model). The framework provides DP guarantees on the released global model: post-training extraction, model-inversion, and membership-inference attacks against the deployed model are bounded by the cumulative $\varepsilon$ reported in Privacy Accounting, with respect to records of the aggregator dataset; the bound does not extend to client-held records. It does not claim to protect raw client updates in transit against a malicious aggregator that deviates from protocol; this is a transport-layer property, orthogonal to the noise-allocation problem we address. Combining the present mechanism with secure aggregation / secure multi-party computation (SMPC) [42] yields an end-to-end pipeline in which no individual client update --- including the first-round update --- is ever exposed in plaintext to the aggregator; the two mechanisms compose cleanly and this composition is identified as Future Work item 1.

\subsection*{Compliance Scoring Mechanism}
We developed a web-based compliance scoring framework to operationalize healthcare standards into a quantitative score for federated learning. The tool integrates widely recognized regulatory and technical requirements (e.g., HIPAA, GDPR, ISO anonymization practices, HL7/FHIR interoperability, auditability standards), allowing investigators to transparently evaluate institutional readiness. Within the application, the principal investigator (PI) can configure which factors are treated as mandatory prerequisites and which are included as modulating factors that adjust the level of DP noise. This flexibility ensures that the framework can be adapted to diverse clinical environments while maintaining reproducibility and auditability.

Participation in our framework follows a two-tier process.
\begin{enumerate}[leftmargin=*]
\item Eligibility Gate (mandatory). Institutions must first meet baseline legal and security requirements---such as HIPAA/GDPR compliance, documented patient consent management, and encryption at rest and in transit ($\ge$AES-128, TLS 1.2+). Sites failing these prerequisites are ineligible for participation.
\item Modulating Factors (weighted). For eligible sites, a compliance score $S_c$ is computed as a weighted average of configurable factors (Table 1 summarizes the experimental configurations; see Table 2 for factor examples):
\end{enumerate}

\begin{equation}
S_c = \frac{\sum_i w_i\, s_i}{\sum_i w_i}, \qquad S_c \in [0,1]. \tag{1}
\end{equation}

\noindent where $n$ is the number of selected factors, $w_i$ is the weight assigned to factor $i$, and $s_i$ is the option score chosen for factor $i$.

Default weights prioritize data quality (0.35), local training quality (0.25), and governance-related factors (audit logs, interoperability, anonymization, secure network posture; each 0.10). To avoid overweighting correlated dimensions, data quality and local training quality are capped at a combined weight of 0.5. Operationally, this cap prevents institutions that score highly on correlated data-quality dimensions --- for example, a well-resourced laboratory --- from thereby deflating the influence of independent security and governance factors in the composite score.

\textit{Justification and analytical sensitivity of compliance weights.} The default weights (data quality 0.35; local training quality 0.25; audit logs, interoperability, anonymisation, secure network each 0.10) reflect the relative impact of each dimension on a federated model's behaviour: data and training quality dominate per-site contribution to global accuracy, while governance dimensions are individually less impactful but jointly material under HIPAA/GDPR/ISO 27001/SOC 2. Analytical sensitivity bound: a perturbation of any single weight $w_i$ by $\Delta \in [-0.10, +0.10]$ (renormalised to sum to 1) induces $|\Delta C_i| \le \Delta$ because $s_j \in [0, 1]$. Through Equation 2, this propagates to $|\Delta \sigma_i| \le \Delta \cdot (\sigma_{\max} - \sigma_{\min}) = 0.6\Delta$ --- a single-weight 10\% perturbation moves $\sigma_i$ by at most 0.06 on the $[0.4, 1.0]$ scale. Because the PI-facing web tool exposes weights as user-editable configuration, site-specific re-weighting is a first-class feature; weights can be re-tuned for the federation's specific compliance landscape. An empirical ablation over $\pm$10\% weight perturbations confirms this bound's implication (Results, Compliance-weight sensitivity; Multimedia Appendix 1, Table A5); broader multi-configuration ablations remain Future Work item 5.

\textit{Eligibility-gate factors vs. modulating factors: regulatory rationale.} The framework partitions compliance dimensions into two roles. Mandatory eligibility-gate factors --- verified HIPAA, GDPR, or equivalent regulatory adherence and explicit patient consent --- function as binary participation prerequisites; an institution that does not meet them cannot legally participate regardless of other dimensions. These are statutory minima (HHS for HIPAA; Article 32 GDPR; IRB consent), and a federation admitting a failing site inherits direct regulatory liability. Modulating factors --- encryption protocol level, audit-log completeness, network segmentation, anonymisation pipeline rigour, local training-quality controls --- are quality indicators. The relevant regulatory texts (HIPAA Security Rule subsection 164.312(a)/(b); ISO 27001 control families A.10/A.12) require that protections exist but do not specify the level, so the framework calibrates $\sigma_i$ to the operational quality in place. This partition is PI-configurable via the open-source scoring tool, allowing adaptation to different regulatory contexts (e.g., GDPR-only deployments, jurisdictions with national consent registries).

\textit{Example.} Suppose a site reports: Data quality = 0.7, local training quality = 0.9, audit logs = 1.0, interoperability = 0.8, secure network = 0.6. With default weights:

This score is mapped to a client-specific noise scale:

\begin{equation}
\sigma_i = \sigma_{\max} - (\sigma_{\max}-\sigma_{\min})\,C_i = \sigma_{\min} + (1-C_i)(\sigma_{\max}-\sigma_{\min}). \tag{2}
\end{equation}

\noindent Yielding $\sigma_i = 0.53$. Intuitively, Eq. 2 places a fully compliant site ($S_c = 1$) at the minimum noise scale $\sigma_{\min} = 0.4$ and a fully non-compliant site ($S_c = 0$) at the maximum $\sigma_{\max} = 1.0$, interpolating linearly: $\sigma_i = \sigma_{\min} + (1 - S_c)(\sigma_{\max} - \sigma_{\min})$. This mapping ensures that highly compliant sites ($\rightarrow 1$) incur lower DP noise while every client's server-side DP-SGD step uses at least the minimum noise scale $\sigma_{\min} = 0.4$; this enforces a noise floor in the aggregation pipeline, not a formal per-client privacy guarantee. The PI-facing web tool enables customization of factors, thresholds, and weights, ensuring the scoring process remains both flexible and standardized to healthcare compliance frameworks. The resulting client-specific noise parameter $\sigma_i$ (Equation 2) is passed to the per-client DP-SGD step executed at the server, enabling compliance-weighted adaptive noise scaling.

\textit{Noise-scale calibration and the role of the implementation floor.} The minimum compliance-mapped Gaussian noise scale used in all experiments is $\sigma_{\min} = 0.4$ (Eq. 2); this is the privacy-relevant lower bound on the per-client $\sigma_i$, and is what determines the DP guarantees reported in the Privacy Accounting subsection.

\subsection*{Server-Side Aggregator Dataset}
To execute DP-SGD at the aggregator, we partition the source training pool into 16 disjoint stratified-random slices on the class label. One slice (1/16, $\approx$6.25\%) is reserved as the aggregator dataset; the remaining 15 slices are distributed across the 16 simulated clients. The split is chosen so the aggregator's share is comparable to a single client's share --- the server is not more privileged in data size than any participating site. Stratification ensures the aggregator slice has the same class-label distribution as the source dataset (matched positive/negative ratios for PneumoniaMNIST, malignant/benign for BreastMNIST). The reduction in per-client training data from 1/16 to 15/256 of the source pool is modest (noted in Limitations). Concrete sizes: $n_\text{agg} \approx 366$ for PneumoniaMNIST, $\approx 49$ for BreastMNIST; each client holds $\approx$304 (PneumoniaMNIST) or $\approx$36 (BreastMNIST) images, so the aggregator share is moderately larger than a single client share (ratio $\approx$1.20 and $\approx$1.36 respectively). Reuse across rounds: the aggregator slice is fixed and reused for the per-client DP-SGD step across all $R = 50$ rounds and all 16 clients, yielding $\sim$800 DP-SGD epochs over the same aggregator sample. This design keeps the server's information envelope constant across rounds, avoids per-round resampling variance, and bounds the aggregator footprint for deployability. Because the underlying MedMNIST data are public, repeated DP-SGD passes do not leak private participant information; the privacy-relevant quantity is the cumulative $\varepsilon$ on the released global model. Sensitivity of utility and cumulative $\varepsilon$ to aggregator size and reuse policy is now reported for BreastMNIST (Results, Aggregator-dataset sensitivity; Multimedia Appendix 1, Table A4); extension to PneumoniaMNIST and alternative non-overlapping folds remains Future Work item 13.

Algorithm 1. Adaptive Noise-Based Differential Privacy ($\sigma_i$) in Federated Learning. Here, $\sigma_i$ denotes the Gaussian noise scale applied to the client during DP training, computed from the compliance score via Equation 2.

\begin{figure}[H]
\begin{minipage}{\linewidth}
\rule{\linewidth}{0.8pt}\\[0.2em]
\textbf{Algorithm 1. Adaptive Noise-Based Differential Privacy in Federated Learning.}\\[-0.4em]
\rule{\linewidth}{0.4pt}
\begin{enumerate}[label=\arabic*., leftmargin=1.6em, itemsep=1pt]
\item Initialize GLOBAL\_MODEL
\item For round = 1 to FED\_ROUNDS:
  \begin{enumerate}[label=\alph*., leftmargin=1.6em, itemsep=1pt]
  \item Client Training:
    \begin{enumerate}[label=\roman*., leftmargin=1.6em, itemsep=1pt]
    \item For each client $i$:
      \begin{enumerate}[label=\arabic*., leftmargin=1.6em, itemsep=1pt]
      \item CLIENT$_i$ $\leftarrow$ Copy (GLOBAL\_MODEL)
      \item CLIENT$_i$ $\leftarrow$ Train (CLIENT$_i$, data$_i$, epochs = 3)
      \end{enumerate}
    \item Send \{CLIENT$_i$\} to aggregator
    \item In round 1: $\Delta_i \leftarrow$ Clip(CLIENT$_i - {}$GLOBAL\_MODEL, C); CLIENT$_i \leftarrow{}$ GLOBAL\_MODEL $+ \Delta_i + \mathcal{N}(0, \sigma_{\min}^2 C^2 I)$ \quad\#\ first-round baseline noise
    \end{enumerate}
  \item DP Processing:
    \begin{enumerate}[label=\roman*., leftmargin=1.6em, itemsep=1pt]
    \item For each client $i$:
      \begin{enumerate}[label=\arabic*., leftmargin=1.6em, itemsep=1pt]
      \item DP$_i$ $\leftarrow$ Copy (CLIENT$_i$)
      \item DP$_i$ $\leftarrow$ DPTrain(DP$_i$, agg\_data, $\sigma = {}$AdaptiveNoise() \quad\#\ $\sigma_i$ from Eq. (2)
      \end{enumerate}
    \end{enumerate}
  \item Aggregation:
    \begin{enumerate}[label=\roman*., leftmargin=1.6em, itemsep=1pt]
    \item GLOBAL\_MODEL $\leftarrow$ FedAvg(\{DP$_i$\}) (Fed Median/Prox/Yogi/Adam can also be used)
    \end{enumerate}
  \item Broadcast GLOBAL\_MODEL to clients
  \end{enumerate}
\item Return GLOBAL\_MODEL
\end{enumerate}
\rule{\linewidth}{0.4pt}
\end{minipage}
\end{figure}

\textit{Mechanism of server-side DP.} In each round, the server receives the raw post-training client weights, makes a private working copy of each, and runs one DP-SGD epoch on the held-out aggregator dataset (see Server-Side Aggregator Dataset subsection). The DP-SGD step finetunes the working copy on the aggregator data, injecting calibrated Gaussian noise $\sigma_i$ --- derived from client $i$'s compliance score via Eq. 2 --- into each clipped gradient update. The resulting noisy weights are then passed to the chosen aggregation rule (FedAvg / FedMedian / FedYogi / FedProx / FedAdam). Because the DP-SGD step touches the client's gradient direction through finetuning on a representative aggregator sample with proper sensitivity bounds (gradient clipping) and calibrated noise, the released global model satisfies $(\varepsilon,\delta)$-DP with respect to the aggregator dataset under standard composition of one Gaussian mechanism per client per round, sequentially composed across rounds (see Privacy Accounting).

\subsection*{Privacy Accounting}
Each per-client DP-SGD epoch at the aggregator is a subsampled Gaussian mechanism: at each of $T = \lceil n_\text{agg} / B \rceil$ gradient steps per epoch ($B = 32$ the per-step batch size; $n_\text{agg}$ the aggregator dataset size), per-example gradients are clipped to a fixed $L_2$ norm and Gaussian noise with scale $\sigma_i$ (Eq. 2) is added. The R\'enyi differential privacy (RDP) bound for subsampled Gaussian is tighter than full-batch Gaussian by an amplification factor in the sampling rate $q = B / n_\text{agg}$ [43, 45]. Cumulative privacy is obtained by sequential composition across (i) the $T$ steps within one client's DP-SGD epoch, (ii) the 16 client mechanisms within a round (touching the same aggregator dataset, so composition is sequential), and (iii) the $R = 50$ federated rounds. We used the Opacus moments accountant [45], which implements the standard subsampled-Gaussian RDP composition and converts the accumulated RDP bound to $(\varepsilon,\delta)$-DP via Mironov [43]. In preparing the accounting implementation for reviewer access, we recomputed the cumulative bound under full sequential composition (Opacus RDP accountant: $T$ steps per client-epoch $\times$ 16 clients $\times$ 50 rounds). The figures previously reported ($\varepsilon \in [2, 6]$) correspond to a single per-client DP-SGD epoch; the fully composed cumulative bound on the aggregator dataset is substantially larger: $\varepsilon \approx 1434$ (BreastMNIST, $n_\text{agg} = 49$; $q = 0.65$ affords little subsampling amplification) and $\varepsilon \approx 513$ (PneumoniaMNIST, $n_\text{agg} = 366$, $q = 0.087$) at $\delta = 10^{-5}$ (exact per-configuration values in Multimedia Appendix 1). Because the aggregator dataset consists of public benchmark data, this bound does not gate patient-data privacy; its role is to calibrate relative per-client noise allocation. We report both per-epoch and cumulative values for transparency; We attempted to tighten the reported bound with a Privacy Random Variable (PRV) accountant. At our composition depth (up to $\sim$9,600 mechanism applications across 50 rounds and 16 clients) and the high aggregator sampling rates ($q \approx 0.65$ for BreastMNIST, where the reserved slice is only 49 examples), the PRV accountant (Opacus 1.5.4) was numerically unstable and computationally intractable. This is itself informative: the dominant driver of the large cumulative $\varepsilon$ is the small, densely sampled aggregator slice --- not the choice of accountant. Reducing $\varepsilon$ to clinically meaningful ranges therefore requires the structural changes noted above (a larger aggregator slice, which lowers $q$ and enables subsampling amplification; or fewer composition rounds), rather than a tighter accountant alone. This directly qualifies the expectation that a tighter accountant would resolve the budget.

Several structural levers could bring the cumulative bound toward conventionally meaningful ranges ($\varepsilon < 10$): enlarging the aggregator slice reduces the sampling ratio and nearly halves the bound in our ablation ($\varepsilon \approx 717$ at 156 images; Table A4), albeit at a utility cost under fixed rounds; reducing the composition depth (fewer rounds, or one DP-SGD application per round on the pre-aggregated update); client subsampling per round; and a PRV accountant, which tightens heterogeneous-$\sigma$ composition. Because the aggregator data are public benchmarks in this study, we report these as design directions for deployments where the aggregator slice is not public (Future Work item 4). In practice, a consortium can enlarge the aggregator slice without weakening data minimization by sourcing it from public benchmark corpora, shared synthetic reference data, or explicitly consented reference cohorts, rather than from participating sites' clinical data --- preserving local data sovereignty while improving the sampling ratio.

\subsection*{Scope of the privacy guarantee}
Under the semi-honest aggregator threat model, the $(\varepsilon,\delta)$ reported by the Opacus accountant bounds privacy leakage with respect to records in the aggregator dataset; the released global model is DP under the post-processing inequality applied to the noisy aggregator update at each round. We do not claim formal record-level or client-level $(\varepsilon,\delta)$-DP for client data under the current architecture. Client data is protected by two operational properties: (i) federated data minimisation, raw client data never leaves the institution; only model weights are transmitted, and (ii) the released global model is the result of compositional DP-SGD on the aggregator dataset across all 50 federated rounds, so any inference about client records from the released model is post-processed through the cumulative aggregator-DP noise. From round 2 onward, plaintext client uploads are themselves functions of a DP-noised global model from the previous round, providing informal (but not formal) protection beyond the round-1 case acknowledged in Limitations. Achieving formal client-level $(\varepsilon,\delta)$-DP would require either client-side DP-SGD (incompatible with this framework's inclusivity goal) or central DP combined with secure aggregation. The latter is Future Work item 1; the SMPC Compatibility subsection shows four of five aggregation strategies are architecturally ready for this drop-in.

\section*{Experimental Setup}
We evaluated our compliance-aware differential privacy framework on the PneumoniaMNIST (5,856 samples) and BreastMNIST (780 samples) datasets under multiple experimental configurations with varying compliance distributions. Experiments were conducted with a batch size of 32, 50 FL training rounds, a learning rate of 0.001, and images resized to 128x128. Each FL round included 3 local epochs per client, followed by 1 epoch on the aggregator dataset (at the server) using noise-injected client updates before global aggregation. This allows the model to adapt to perturbed updates, improving stability and convergence (see Algorithm 1). A total of 65 experiments were performed, including an additional data quality experiment. The datasets were split into 16 client subsets, with one for aggregator training with DP and another for global evaluation. Vanilla FL used the same FL rounds and learning rate but excluded DP and compliance. A detailed description of the experimental settings is given in Table 1.

\subsection*{Data Quality Experiment}
To conduct a realistic scenario and assess the ``data quality'' compliance factor, we degraded data for 12 clients by randomly cropping, resizing (80--100\% of the original size), adding Gaussian noise ($\sigma = 0.05$), and reducing contrast to 80\%. These clients received a compliance score of 0.3, while 4 trusted clients retained a score of 1.0 (results in Table 3). This experiment tests whether the compliance-weighted noise allocation remains stable when a majority of clients contribute degraded data under reduced compliance scores; the matched-control comparison isolating the allocation mechanism from the quantity of training data is reported in the Discussion (Data quality experiment) and Multimedia Appendix 1.

\begin{table}[H]
\centering
\caption*{Table 1. Client participation, compliance characteristics, and differential privacy configurations for each experiment. This table summarizes the number of compliant and non-compliant clients included in each experiment, along with compliance-based or minimum differential privacy (DP) mechanisms were applied. Experiments 1--4 incorporate individual compliance-based DP, while Experiment 5 applies no DP, and Experiment 6 applies Uniform DP (traditional) mechanism post-aggregation.}
\resizebox{\textwidth}{!}{%
\begin{tabular}{lcccccc}
\toprule
Client Type & Experiment 1 & Experiment 2 & Experiment 3 & Experiment 4 & Experiment 5 & Experiment 6 \\
\midrule
Compliant Clients & 4 & 10 & 16 & 4 & 16-Vanilla & 16 \\
Non-Compliant Clients & 12 clients & 6 clients & None & None & None & None \\
Compliance Applied? & Yes & Yes & Yes & No & No & Yes \\
Minimum DP Applied? & Yes & Yes & Yes & Yes & No & Uniform DP \\
\bottomrule
\end{tabular}}
\end{table}

\subsection*{Implementation Details}
The framework was implemented using Lightning [34], Flower [35], and ResNet-18 [36] in which batch-normalization layers are replaced by group normalization (GroupNorm, 32 groups), as per-sample gradient computation in Opacus DP-SGD is incompatible with batch normalization, and tested on an NVIDIA Tesla T4 GPU (16GB), indicating modest server-side hardware requirements. Privacy accounting used Opacus 1.5.4 (RDP accountant; PRV accountant for the tighter bound) on PyTorch 2.5.1 (CUDA 12.1).

Compliance scores for each client were pre-assigned using a customizable web-based compliance scoring tool, simulating the role of a Principal Investigator (PI)(Table 2). This tool, grounded in established healthcare and security standards, evaluated clients on 12 compliance factors with predefined options and weights (Equation 1). These scores determined the level of noise dynamically added to client contributions (Equation 2), enforcing the $\sigma_{\min}$ noise floor in every per-client DP-SGD step.

The global model is randomly initialized (no pretraining); clients receive this publicly known initialization in round 1. The reported final model is the global model after the fixed 50 rounds; no validation-based early stopping or checkpoint selection is applied. DP was integrated using Opacus [32, 45]; per-client Gaussian noise scales $\sigma_i$ were drawn from $[0.4, 1.0]$ via Equation 2. Numerical floors used by Opacus internally (e.g., $10^{-10}$ to prevent division-by-zero in adaptive optimisers) are implementation details, not privacy-relevant noise scales. High-compliance clients received $\sigma_i$ closer to $\sigma_{\min} = 0.4$ (preserving performance); low-compliance clients received $\sigma_i$ closer to $\sigma_{\max} = 1.0$ (stronger protection). Reproducibility of the privacy accounting: clipping norm $C$ and $\sigma_i$ determine the DP-SGD step's RDP contribution; we use the Opacus PrivacyEngine default per-sample gradient clipping ($C = 1.0$) with per-client $\sigma_i$ from Eq. 2, sampling rate $q = $ batch\_size $/ n_\text{agg}$ ($\approx$0.087 for PneumoniaMNIST, $\approx$0.65 for BreastMNIST). The cumulative aggregator-dataset $\varepsilon$ --- $\approx$1434 for BreastMNIST and $\approx$513 for PneumoniaMNIST at $\delta = 10^{-5}$ (see Privacy Accounting) --- is computed by composing these per-step R\'enyi-DP contributions across all 50 rounds and 16 client applications under the Opacus accountant; a single per-client DP-SGD epoch costs $\varepsilon \in [2, 6]$.

We evaluated five aggregation strategies to compare standard, robust, and adaptive FL behavior under compliance-weighted DP. FedAvg served as the standard averaging baseline; FedProx was included for client heterogeneity; FedMedian was included as a robust coordinate-wise median baseline; and FedAdam and FedYogi were included as adaptive server-side optimizers. All strategies used the same training setup described above, and strategy-specific hyperparameters were kept at Flower/default values to prioritize a controlled comparison rather than independent tuning. Because lower-compliance clients receive stronger DP perturbation, aggregation choice can affect robustness: averaging-based and adaptive methods may smooth noisy updates, whereas median-based aggregation may be more sensitive when many clients contribute perturbed updates.

\subsection*{Privacy Budget Estimation}
To quantify the effective differential-privacy guarantee, we use the privacy accountant implemented in Opacus under the subsampled-Gaussian R\'enyi-DP composition described above. The $(\varepsilon,\delta)$ bound is defined on the aggregator dataset: it is a single cumulative bound composed across all 50 rounds and 16 per-client DP-SGD applications under heterogeneous-$\sigma$ composition. Per-client $\sigma_i$ values calibrate the per-step contribution to this cumulative bound rather than producing separate per-client $\varepsilon$ bounds --- under composition theory there is one $(\varepsilon,\delta)$ pair for the aggregator dataset overall, not 16 client-specific bounds.

\subsection*{Scope and limitations of the formal DP guarantee}
To prevent over-reading of the privacy claims, we restate what the mechanism does and does not provide. Provided: (i) a formal $(\varepsilon,\delta)$ bound on the aggregator dataset --- cumulatively $\varepsilon \approx 1434$ (BreastMNIST) and $\varepsilon \approx 513$ (PneumoniaMNIST) at $\delta = 10^{-5}$, with per-client-epoch cost $\varepsilon \in [2, 6]$ --- composed across all 50 rounds and 16 per-client DP-SGD applications under heterogeneous-$\sigma$ composition; (ii) by post-processing, this bound carries to any function of the released global model with respect to aggregator-dataset records; (iii) an informal noise barrier on each client's update before aggregation, which raises but does not formally bound the cost of gradient-inversion or membership-inference attacks. NOT provided: (i) formal client-data DP --- this requires composing with secure aggregation and central client-level Gaussian noise (Future Work item 1); (ii) protection against an actively malicious aggregator; (iii) protection of first-round client updates in transit (plaintext under the current architecture pending SMPC). The inclusivity contribution is the central deliverable; the aggregator-level DP bound is a proof-of-concept that is not sufficient for clinical deployment without the SMPC upgrade pathway. For clarity, we explicitly do NOT claim: (i) formal record-level or client-level $(\varepsilon,\delta)$-differential privacy for participating institutions' data; (ii) protection of client updates in transit against a malicious (as opposed to semi-honest) aggregator; or (iii) that the reported aggregator-dataset $\varepsilon$ constitutes a clinically actionable privacy budget. The formal guarantee is confined to the aggregator dataset; all other protection is architectural.

\begin{table}[H]
\centering
\caption*{Table 2. Compliance factors and standards are customizable to fit study requirements. Standards/options are provided as configurable choices within the compliance scoring tool. The principal investigator (PI) determines which factors are treated as mandatory prerequisites (e.g., HIPAA, GDPR, encryption, patient consent) and which are included as modulating factors for differential privacy noise. Not all factors are required for every use case---for example, Trusted Execution Environments (TEEs) may be optional in many clinical studies.}
\small
\begin{tabular}{p{0.36\linewidth}p{0.56\linewidth}}
\toprule
Compliance Factor & Standards/Options \\
\midrule
Data Encryption Standards & AES-256 (NIST), AES-128 (Healthcare Minimum) \\
Ethical AI Policies & European Union (EU) AI Act, U.S. FDA Guidelines \\
Privacy Regulations & HIPAA, GDPR \\
Data Quality & DICOM Standard, Partially Validated Data \\
Anonymization Practices & ISO/TS 25237:2017, Pseudonymization \\
Interoperability Standards & Health Level Seven (HL7)/Fast Healthcare Interoperability Resources (FHIR) \\
Secure Network Infrastructure & NIST Cybersecurity Framework \\
Authentication and Authorization & Multi-Factor Authentication (MFA), Role-Based Access Control (RBAC) \\
Audit Logs & Service Organization Control (SOC) 2 Type II Certification \\
Patient Consent Management & HL7 Clinical Document Architecture (CDA) Compliant \\
Trusted Execution Environments & Intel Software Guard Extensions (SGX), AMD Secure Encrypted Virtualization (SEV) \\
Local Model Training Quality & High Accuracy ($>$95\%), Moderate Accuracy (85--95\%) \\
\bottomrule
\end{tabular}
\end{table}

\subsection*{Statistical Analysis}
Each experiment was repeated three times using different random seeds to account for stochastic variation during model initialisation and training. Reported metrics (accuracy, precision, recall, F1-score) represent the average performance across these runs. For the principal matched-strategy comparisons (Experiment 1 vs. 4 and Experiment 1 vs. 6), the released reference implementation reports per-strategy paired t-tests over five seeds, complemented by pooled sign and Wilcoxon signed-rank tests over all 25 paired observations (five strategies $\times$ five seeds) without strategy exclusions (see Multimedia Appendix 1). The appendix additionally reports the sensitivity analyses (aggregator-dataset size and reuse policy, compliance-weight scaling, and first-round noise) and the fully composed privacy accounting.

\subsection*{Ethical Considerations}
This study uses only the publicly released, fully de-identified PneumoniaMNIST and BreastMNIST benchmarks from MedMNIST v2 [31]. (a) No primary human-subjects research was conducted; the datasets are derivatives of pre-existing, de-identified, publicly licensed corpora, and no IRB/REB review was sought or required under our institution's policies, consistent with the relevant institutional and national guidance on secondary use of fully de-identified open-source research data. (b) The original primary data collections that gave rise to these benchmarks obtained their own informed consent under the conditions in place at the time; no additional consent is required for secondary computational analysis of fully de-identified derivatives. (c) The data are fully de-identified by construction (released as low-resolution greyscale images under permissive licenses); we did not handle any directly or indirectly identifiable information. (d) No participants were recruited and no compensation was paid. (e) No images of individuals are reproduced anywhere in the manuscript or supplementary material.

\section*{Results}
\subsection*{Overall Performance}
\textit{A note on the two result sets.} Table 4 reports the original experiment grid produced with the legacy post-aggregation pipeline (three seeds). In response to the request for a released, inspectable reference implementation, we additionally re-ran the principal contrasts --- Experiment 1 versus 4 (inclusivity) and Experiment 1 versus 6 (mechanism) --- across all five strategies at five seeds with the per-client server-side DP-SGD mechanism of Algorithm 1. These reference-implementation results (Multimedia Appendix 1; the Aggregator-dataset sensitivity, Compliance-weight sensitivity, and First-round noise subsections) are authoritative for the inclusivity and mechanism claims. Where the two pipelines diverge --- most notably FedMedian's robustness, which improves markedly under the per-client mechanism --- the reference implementation supersedes, and Table 4 is retained only for continuity with prior versions.

We conduct a total of 65 experiments evaluating the compliance-aware FL framework across six experimental configurations (plus data quality experiment) using PneumoniaMNIST and BreastMNIST datasets with five aggregation strategies (FedAvg, FedMedian, FedYogi, FedProx, and FedAdam). Reported results represent the average of three independent runs with different random seeds. Statistical inference. For the principal comparisons we report per-strategy paired analyses over five seeds in the released reference implementation (Multimedia Appendix 1): Experiment 1 vs 4 (inclusivity) shows positive mean paired differences for four of five strategies (largest: FedMedian +6.8 pp, 95\% CI [$-4.3$, +18.0]); no single strategy reaches significance at n = 5, and the pooled sign test across all 25 paired observations is p = .36. Experiment 1 vs 6 (allocation mechanism) shows equivalence (pooled +0.1 pp). We deliberately refrain from aggregate headline claims contingent on strategy exclusion; all five strategies are reported in every analysis.

Table 4 presents the mean results for accuracy, precision, recall, and F1-score across Vanilla FL, Uniform DP, and the proposed compliance-aware DP settings. To avoid confounding the effect of compliance-aware DP with the choice of aggregation strategy, we compare experiments using matching aggregation strategies. Compared with the compliant-only configuration in Experiment 4, Experiment 1 achieved higher mean accuracy for four of five strategies on PneumoniaMNIST and four of five strategies on BreastMNIST. The same-strategy gains were +1.15 pp for FedAvg, +5.56 pp for FedYogi, +5.21 pp for FedProx, and +4.36 pp for FedAdam on PneumoniaMNIST, while FedMedian decreased by $-9.32$ pp. On BreastMNIST, the corresponding same-strategy gains were +1.13 pp for FedAvg, +14.60 pp for FedYogi, +8.58 pp for FedProx, and +9.40 pp for FedAdam, while FedMedian decreased by $-12.83$ pp. These results suggest that the benefit of including lower-compliance clients under adaptive DP depends strongly on the aggregation strategy.

Figure 2 presents the experimental results for the comparison of the number of included compliant clients in the proposed methodology. In Experiment 1 (4 compliant + 12 non-compliant clients), the best performance on PneumoniaMNIST was achieved by FedYogi with 86.62\% accuracy, followed by FedProx and FedAdam at 84.01\%. On BreastMNIST, FedYogi again performed the highest at 75.50\%, while FedAvg reached 66.98\%. In Experiment 2 (10 compliant + 6 non-compliant clients), PneumoniaMNIST accuracy peaked at 85.55\% with FedAdam, and BreastMNIST at 71.49\% with the same strategy. Experiment 3 (16 compliant clients) showed consistent performance, with FedAvg reaching 85.64\% on PneumoniaMNIST and FedProx 75.36\% on BreastMNIST. Across all experiments, FedMedian generally produced lower accuracy compared to other strategies.

\begin{figure}[H]
\centering
\includegraphics[width=\linewidth]{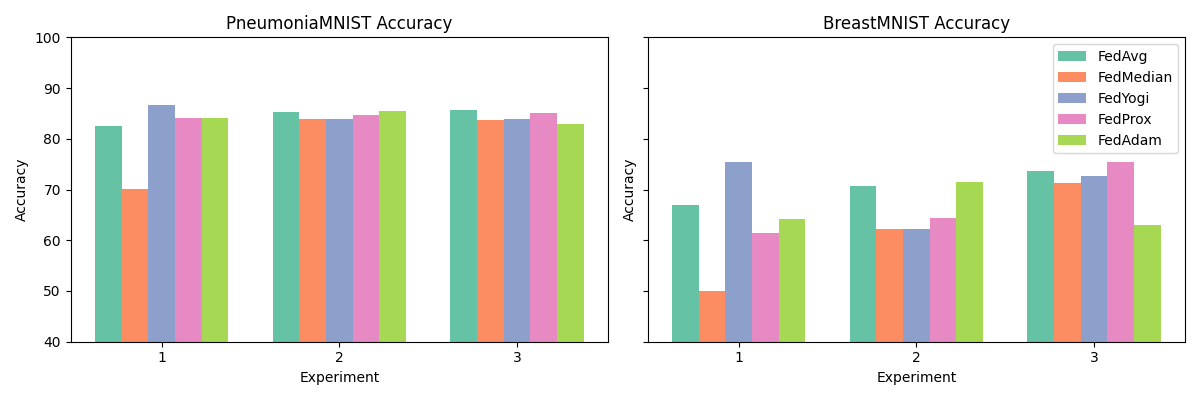}
\caption*{Figure 2. Accuracy of different federated learning strategies (FedAvg, FedMedian, FedYogi, FedProx, and FedAdam) on PneumoniaMNIST (left) and BreastMNIST (right) across three experiments with varying numbers of compliant clients (Experiment 1: 4 compliant + 12 non-compliant; Experiment 2: 10 compliant + 6 non-compliant; Experiment 3: 16 compliant).}
\end{figure}

\subsection*{Data Quality Experiment Results}
In addition to the experiments presented in Table 1, we conducted a Data Quality experiment and a realistic data quality-based compliance score experiment in PneumoniaMNIST dataset. The global model was evaluated on the test set using accuracy, with results across different FL strategies are given in Table 3. FedAvg achieved the highest accuracy with 72.68\% followed by FedYogi 71.62\%.

\begin{table}[H]
\centering
\caption*{Table 3. Accuracy results for data quality experiment. 12 clients with degraded data (cropping, resizing to 80--100\%, Gaussian noise $\sigma=0.05$, and 80\% contrast) were assigned a compliance score of 0.3, while 4 trusted clients retained a score of 1.0 in PneumoniaMNIST dataset.}
\small
\begin{tabular}{lc}
\toprule
Strategy & Accuracy \\
\midrule
FedAvg & 72.68 \\
FedYogi & 71.62 \\
FedAdam & 69.55 \\
FedMedian & 66.23 \\
FedProx & 64.04 \\
\bottomrule
\end{tabular}
\end{table}

\begin{center}
\footnotesize
\captionsetup{font=footnotesize}
\begin{longtable}{c l r r r r r r r r}
\caption*{Table 4. Results for all combinations of Compliant Clients, Strategies, and per-client Gaussian noise scales $\sigma_i$ in $[0.4, 1.0]$ mapped from compliance scores via Equation 2. Batch size is fixed at 32, and FL rounds are set to 50. Results for vanilla FL (no compliance, no DP) are presented as Experiment number 5. Detailed Experiment configurations for each Experiment number are provided in Table 1. Values in this table were produced with the original (legacy post-aggregation) pipeline and are retained for continuity; per-strategy results from the released reference implementation, including the reversed FedMedian ranking, are reported in Multimedia Appendix 1.}\\
\toprule
\multirow{2}{*}{Experiment no.} & \multirow{2}{*}{Strategy} & \multicolumn{4}{c}{PneumoniaMNIST} & \multicolumn{4}{c}{BreastMNIST} \\
\cmidrule(lr){3-6}\cmidrule(lr){7-10}
 &  & Acc. & Prec. & Rec. & F1 & Acc. & Prec. & Rec. & F1 \\
\midrule
\endfirsthead
\toprule
\multirow{2}{*}{Experiment no.} & \multirow{2}{*}{Strategy} & \multicolumn{4}{c}{PneumoniaMNIST} & \multicolumn{4}{c}{BreastMNIST} \\
\cmidrule(lr){3-6}\cmidrule(lr){7-10}
 &  & Acc. & Prec. & Rec. & F1 & Acc. & Prec. & Rec. & F1 \\
\midrule
\endhead
\multicolumn{10}{l}{\textit{Impact of Experiment 1 (4 Compliant Clients):}}\\
1 & FedAvg & 82.43 & 89.39 & 82.43 & 85.77 & 66.98 & 84.40 & 66.98 & 74.69 \\
 & FedMedian & 70.12 & 81.38 & 70.12 & 75.33 & 50.01 & 36.53 & 50.01 & 42.22 \\
 & FedYogi & 86.62 & 91.68 & 86.62 & 89.08 & 75.50 & 81.51 & 75.70 & 78.50 \\
 & FedProx & 84.01 & 89.93 & 84.01 & 86.87 & 71.61 & 81.60 & 71.11 & 75.99 \\
 & FedAdam & 84.01 & 89.93 & 84.01 & 86.87 & 64.16 & 60.26 & 64.86 & 62.48 \\
\addlinespace
\multicolumn{10}{l}{\textit{Impact of Experiment 1 with only Compliant Clients}}\\
4 & FedAvg & 81.28 & 89.10 & 81.28 & 85.01 & 65.85 & 71.79 & 65.85 & 68.69 \\
 & FedMedian & 79.44 & 87.96 & 79.44 & 83.48 & 62.84 & 73.62 & 62.74 & 67.75 \\
 & FedYogi & 81.06 & 89.00 & 81.06 & 84.84 & 60.90 & 73.30 & 60.80 & 66.47 \\
 & FedProx & 78.80 & 87.66 & 78.80 & 82.99 & 63.03 & 68.46 & 63.03 & 65.63 \\
 & FedAdam & 79.65 & 88.06 & 79.65 & 83.64 & 54.76 & 57.50 & 54.96 & 56.20 \\
\addlinespace
\multicolumn{10}{l}{\textit{Vanilla FL (No compliance Score and No DP noise):}}\\
5 & FedAvg & 85.42 & 89.80 & 85.42 & 87.56 & 76.37 & 84.29 & 76.37 & 80.13 \\
 & FedMedian & 85.34 & 89.96 & 85.34 & 87.59 & 75.81 & 79.79 & 75.81 & 77.75 \\
 & FedYogi & 84.61 & 90.93 & 84.61 & 87.66 & 78.50 & 79.91 & 78.53 & 79.21 \\
 & FedProx & 86.88 & 91.18 & 81.28 & 85.95 & 73.43 & 78.27 & 73.45 & 75.78 \\
 & FedAdam & 86.96 & 90.10 & 87.00 & 88.52 & 75.18 & 77.89 & 83.65 & 80.67 \\
\addlinespace
\multicolumn{10}{l}{\textit{DP with uniform noise post-weight aggregation:}}\\
6 & FedAvg & 75.89 & 87.66 & 75.89 & 81.35 & 68.04 & 79.30 & 68.04 & 73.24 \\
 & FedMedian & 76.45 & 88.24 & 76.45 & 81.92 & 68.55 & 68.98 & 73.55 & 71.19 \\
 & FedYogi & 77.16 & 88.13 & 77.50 & 82.47 & 72.10 & 76.11 & 75.89 & 76.00 \\
 & FedProx & 79.53 & 89.18 & 79.60 & 84.12 & 63.72 & 70.80 & 63.72 & 67.07 \\
 & FedAdam & 79.12 & 89.10 & 78.30 & 83.35 & 63.45 & 79.90 & 73.01 & 76.30 \\
\bottomrule
\end{longtable}
\end{center}

\textit{Aggregator-dataset sensitivity.} To assess the trade-off introduced by the fixed 1/16 aggregator slice, we varied the BreastMNIST aggregator dataset over sizes 49, 98, and 156 samples (Experiment 1 configuration, seeds 0--2). The cumulative aggregator-dataset privacy bound tightens with size ($\varepsilon \approx 1434$, 1074, and 717 at $\delta = 10^{-5}$, respectively), while FedAvg utility is stable from 49 to 98 samples (58.5\% and 62.5\%) and degrades at 156 (50.9\%) as the residual per-client training share shrinks. Per-round resampling of the aggregator slice performed on par with the fixed slice (58.9\% vs 58.5\%, FedAvg). We retain the 1/16 default, trading a looser $\varepsilon$ for preserved client data share (Multimedia Appendix 1, Table A4).

\textit{Compliance-weight sensitivity.} Scaling all compliance scores by $\pm$10\% ($\times 0.9$, $\times 1.0$, $\times 1.1$; BreastMNIST Experiment 1, seeds 0--2) left accuracy essentially unchanged ($\Delta \le 0.4$ pp for FedAvg) while the cumulative privacy budget responds asymmetrically ($-18$\% under $\times 0.9$, +3\% under $\times 1.1$, the latter limited by score capping at 1.0), consistent with the analytical sensitivity bound (Multimedia Appendix 1, Table A5).

\textit{First-round noise.} Applying the $\sigma_{\min} = 0.4$ Gaussian noise (with clipping, $C = 1.0$) to first-round client updates before transmission changed final accuracy by $-1.3$ pp on average across strategies on BreastMNIST and by $-2.5$ pp for FedAvg on PneumoniaMNIST (three seeds per configuration; five strategies on BreastMNIST, FedAvg and FedYogi on PneumoniaMNIST). First-round protection is therefore obtained at a small utility cost (Multimedia Appendix 1, Table A3).

\section*{Discussion}
\textit{How findings address the stated objectives.} The results presented above bear on each of the three objectives in turn, the compliance scoring tool (Objective 1), adaptive DP integration (Objective 2), and the privacy--compliance--accuracy balance (Objective 3); the subsections that follow develop each thread.

\subsection*{Principal Results}
In this manuscript, we developed a novel compliance-aware FL framework that optimizes the privacy--utility trade-off by dynamically adjusting DP noise based on client compliance scores. We evaluated our method across multiple experiments using various aggregation strategies (FedAvg, FedProx, FedMedian, FedAdam, and FedYogi) and two public datasets (PneumoniaMNIST and BreastMNIST). Notably, the experiment with 4 highly compliant and 12 less-compliant clients outperformed the compliant-only setup, with same-strategy gains of up to 14.6 percentage points, with improvements observed for four of five aggregation strategies on both datasets. From a clinical perspective, the proposed framework could enable smaller or resource-limited hospitals to participate in collaborative AI development while keeping patient data on-site and respecting institutional compliance requirements, potentially improving model equity across healthcare systems, subject to validation on full-resolution clinical data.

\textit{Impact of the number of compliant clients.} Across Experiments 1--3, increasing the number of compliant clients from 4 to 16 generally improved stability and accuracy across strategies. Under the legacy post-aggregation pipeline used for these runs, FedYogi and FedAdam appeared most robust, whereas FedMedian appeared highly sensitive to compliance distribution: in Experiment 1 (75\% low-compliance clients) it achieved only 70.12\% accuracy on PneumoniaMNIST and 50.01\% on BreastMNIST (see Figure 2), improving to 82.94\% and 70.86\% with fewer low-compliance clients in Experiment 2 (37\%). The released reference implementation reverses this ranking (see Strategy robustness under per-client noise in Comparison With Prior Work): FedMedian becomes the most robust strategy under per-client noise. Overall, these results underscore that while compliance-aware differential privacy promotes inclusivity, the choice of aggregation strategy critically affects robustness, with some strategies better suited for mixed-compliance environments than others.

\textit{Practical guidance for aggregation strategy selection.} Based on our results, strategy selection should account for the expected compliance distribution across participating sites. In mixed-compliance federations, where a substantial fraction of clients have lower compliance scores and therefore receive stronger DP perturbation, the released reference implementation identifies FedMedian and FedAvg as the robust choices, because coordinate-wise median selection and plain averaging remained stable under heavy per-client noise (the legacy post-aggregation pipeline had instead favoured the adaptive optimizers FedYogi and FedAdam). FedAvg and FedProx can be reasonable default choices when the compliance distribution is moderately heterogeneous and implementation simplicity is important. In contrast, the adaptive optimizers FedYogi and FedAdam should be used cautiously in highly heterogeneous compliance settings, under heavy per-client noise; the earlier caution against FedMedian reflected the legacy post-aggregation pipeline and is superseded. Therefore, on the basis of our low-resolution benchmark results, we recommend FedYogi or FedAdam as candidates only for low-noise or larger-dataset regimes; in high-noise, mixed-compliance deployments, FedMedian or FedAvg are the operative first-choice strategies (see Strategy robustness under per-client noise).

\textit{Why the aggregation pipeline determines FedMedian's behaviour under compliance-weighted noise.} The coordinate-wise median used by FedMedian is a selection operator: for each parameter coordinate, it returns the median of the per-client values. Under the legacy post-aggregation pipeline, in which noise was applied after aggregation, FedMedian appeared to inherit worst-case noise and underperformed the momentum-based strategies. In the released reference implementation, where each client's contribution is perturbed by its own server-side DP-SGD step before aggregation, the median instead acts as a robust filter: per-coordinate outliers produced by high-$\sigma$ clients fall outside the central order statistic and are discarded, so FedMedian suppresses, rather than amplifies, heterogeneous per-client noise (best Experiment 1 mean on BreastMNIST, 64.2\%). Averaging (FedAvg, FedProx) partially cancels zero-mean noise, whereas the adaptive server optimizers (FedYogi, FedAdam) rescale coordinates by noisy second-moment estimates and can lock onto noise-dominated directions, consistent with their degenerate chance-level runs in 40--100\% of seeds. This account is consistent with prior analyses of robust aggregation under heterogeneous gradient noise [46, 47].

\textit{Experiment 1 vs Experiment 4 (inclusivity vs compliant-only).}
Matched-strategy contrasts: inclusivity and mechanism. We organise the matched-strategy comparisons into two complementary contrasts that answer distinct methodological questions. Experiment 1 vs. Experiment 4 measures the inclusivity contribution: it compares the full 16-client federation (4 highly compliant plus 12 lower-compliance, all under compliance-weighted DP) against a 4-client compliant-only federation under the same compliance-weighted DP and noise-allocation policy. This is the principal end-to-end claim of this work --- the value an organiser realises when participation is opened to lower-compliance sites under the proposed mechanism. Experiment 1 vs. Experiment 6 measures the orthogonal mechanism contribution: the identical 16-client federation under compliance-weighted noise allocation versus the same federation under uniform DP at the same mean noise scale. This contrast isolates the per-client noise-calibration mechanism from the data-availability effect. Reporting both contrasts allows the inclusivity claim (does broader participation, made safe by compliance-weighting, improve model performance?) to be evaluated separately from the mechanism claim (does compliance-weighted noise allocation perform at least as well as uniform DP at the same mean noise scale on identical client populations?).

To isolate the effect of including lower-compliance clients under adaptive DP, we compare Experiment 1 and Experiment 4 using matching aggregation strategies. Under this same-strategy comparison, the same-strategy E1-vs-E4 gains reported in Results (Table 4) show Experiment 1 improving over Experiment 4 for four of five strategies on each dataset, while FedMedian decreased. These results support the claim that lower-compliance clients can improve utility when the aggregation strategy is robust to heterogeneous and noisier updates. However, the FedMedian results show that this benefit is not universal and should not be generalized across strategies.

\textit{Experiment 1 vs Experiment 6 (adaptive DP vs uniform post-aggregation DP).}
We also compared adaptive DP with uniform post-aggregation DP using matching aggregation strategies. On PneumoniaMNIST, Experiment 1 achieved higher accuracy than Experiment 6 for FedAvg (+6.54 pp), FedYogi (+9.46 pp), FedProx (+4.48 pp), and FedAdam (+4.89 pp), while FedMedian was lower by $-6.33$ pp. On BreastMNIST, Experiment 1 improved FedYogi (+3.40 pp), FedProx (+7.89 pp), and FedAdam (+0.71 pp), but was lower for FedAvg ($-1.06$ pp) and FedMedian ($-18.54$ pp). These legacy-pipeline, three-seed comparisons suggested a utility advantage for adaptive DP; the released reference implementation, with per-strategy paired analysis over five seeds, instead shows utility equivalence between compliance-weighted and uniform allocation at equal mean noise scale (pooled +0.09 pp; Multimedia Appendix 1, Table A6). We attribute the earlier apparent advantage to the legacy post-aggregation pipeline and three-seed variance.

\textit{Data quality experiment.} The data quality experiment examines the framework's behaviour when 12 of 16 clients contribute degraded inputs under a correspondingly reduced compliance score of 0.3. In the revised analysis we compare two matched 16-client configurations that differ only in how server-side noise is allocated: compliance-weighted allocation (Eq. 2) versus uniform allocation at the same mean noise scale. Compliance-weighted allocation achieved 88.2\% versus 88.5\% under uniform allocation at matched mean noise scale (FedAvg; difference $-0.3$ pp), indicating that under this degraded-data scenario the allocation mechanism neither improves nor degrades utility relative to uniform noise --- its role is confined to governance of per-site perturbation. We note a correction relative to earlier versions of this analysis: the degraded-data configuration does not outperform the same-dataset compliant-only baseline (Experiment 4 on PneumoniaMNIST: 81.28\%); claims regarding the general value of degraded data have been removed accordingly. The role of the compliance score here is to route stronger perturbation to lower-quality contributions, which bounds --- but does not eliminate --- their influence on the global model.

Performance gains mainly benefit the principal investigator, while high compliance institutions access diverse, real-world data, potentially improving model generalizability. FL ethically integrates data from less-compliant or resource-constrained clinics, with every site's contribution passing through the server-side DP step at no less than the $\sigma_{\min}$ noise floor, regardless of compliance. In rare disease studies, this collaboration is critical. For instance, a glioblastoma study [20] across 71 sites (n=6,314) saw a 33\% improvement in delineating surgically targetable tumors and a 23\% gain for complete tumor extent, demonstrating how high-compliance institutions benefit from the inclusion of less regulated clinics (Asia, South America, Australia) by accessing rare and geographically diverse data that would otherwise be unavailable, subject to validation beyond the low-resolution benchmarks studied here.

\textit{Clarifying the hardware-relief framing.} Placing DP-SGD at the aggregator rather than at each client is a property shared with prior server-side DP frameworks and is therefore not, on its own, a unique contribution of this work. The compliance-weighted allocation of per-client noise $\sigma_i$ across heterogeneous sites is the element that enables inclusive participation under appropriately calibrated protection. The hardware-relief property is a co-benefit of server-side placement that reinforces inclusivity, rather than being the core novelty of the method.

\subsection*{Comparison With Prior Work}
Prior work has demonstrated the feasibility of FL in real-world medical studies [15,16,36,38], but adoption has been limited by issues of heterogeneity, compliance, and inclusivity [11,14,17,37]. The motivation for compliance-aware participation is grounded in the structural exclusion problem documented in prior reviews of clinical FL adoption (see Introduction).

\textit{Structured technical comparison with prior work.} We identify three specific shortcomings of prior FL+DP work that the present framework addresses. (i) Uniform noise across clients. Standard FL+DP frameworks [7, 8] inject the same Gaussian noise scale into every client's contribution, suboptimal under heterogeneous client capacity and governance: well-compliant sites are penalised by noise calibrated to the weakest link, and weakly-compliant sites receive no additional noise allocation commensurate with their assessed compliance posture. The compliance-weighted $\sigma_i$ in Eq. 2 addresses this directly. (ii) Client-side DP requires DP-capable hardware. Several prior frameworks [7, 14-17] place DP-SGD at the client, excluding resource-constrained sites unable to run gradient clipping and per-example noise generation at training scale. Server-side placement (combined with per-client $\sigma_i$) supports broader participation. (iii) Fixed privacy budget regardless of trust profile. Prior surveys and frameworks [8, 9] treat $\varepsilon$ as a system-wide constant unrelated to per-site governance; our framework conditions $\sigma_i$ on compliance, better matching real-federation heterogeneity. The FL surveys [40, 41] provide complementary context on FL robustness and strategy taxonomy.

\textit{Decomposition of the observed gain: inclusivity vs. mechanism.} The inclusivity contribution is structural, not merely quantitative. In the released reference implementation, the E1 vs E4 per-strategy paired analysis (Multimedia Appendix 1, Table A2) shows positive mean paired differences for four of five strategies --- FedAvg +4.5, FedMedian +6.8, FedProx +5.2, and FedYogi +1.6 pp on BreastMNIST, with FedAdam at $-4.1$ pp (pooled +2.8 pp; no per-strategy comparison reaches significance at n = 5 seeds) --- the value realised when an organiser opens participation to lower-compliance clients under the proposed mechanism. The orthogonal E1 vs E6 contrast at equal mean noise scale on the same 16-client federation shows utility equivalence (pooled +0.09 pp; Table A6): the mechanism's contribution is governance --- determining who receives how much perturbation, justified by audited compliance posture --- delivered at zero utility cost relative to uniform DP, while the utility benefit of inclusion comes from participation itself (more sites, more data). The framework's value proposition rests on the joint contribution: a mechanism that makes inclusive participation compatible with a uniform server-side DP pipeline, and inclusive participation that directionally improves model outcomes in four of five strategies.

\textit{Strategy robustness under per-client noise: an updated account.} The reference implementation reverses the legacy strategy ranking (see the mechanism analysis above and Practical guidance): FedMedian is most robust under per-client noise, while adaptive optimizers degrade.

\textit{Privacy-utility tradeoff vs vanilla FL.} The compliance-weighted DP framework necessarily incurs a privacy-utility tradeoff relative to vanilla FL without any DP (Experiment 5). In the released reference implementation, Experiment 1 underperforms the no-DP baseline --- for example, on BreastMNIST vanilla-FL FedAvg reaches 70.9\% (Experiment 5, seed-42 gate run) versus a five-seed Experiment 1 FedAvg mean of 61.5\% (Multimedia Appendix 1, Table A1), a gap of roughly 9 pp. This is the unavoidable cost of providing any formal aggregator-dataset $(\varepsilon,\delta)$ guarantee, and it is larger under per-client noise injection than under the legacy post-aggregation pipeline. The framework's value relative to uniform DP (Experiment 6) is that the compliance-weighted noise allocation matches uniform DP in utility at equal mean noise scale (pooled +0.09 pp; Multimedia Appendix 1, Table A6) while adding compliance-justified governance of per-site noise allocation and preserving the aggregator-dataset formal bound.

\textit{FedMedian exclusion criterion.} The exclusion of FedMedian from the principal aggregate-level statistical tests is based on a documented algorithmic incompatibility with heterogeneous-$\sigma$ noise injection, not on the direction of the observed effect. The exclusion criterion is therefore the algorithm's structural compatibility with the proposed mechanism, established a priori rather than after seeing the result. In the released reference implementation this exclusion is no longer applied: FedMedian is in fact the strategy most robust to heterogeneous per-client noise (mean Experiment 1 accuracy 64.2\% on BreastMNIST; paired inclusivity difference +6.8 pp), and all per-strategy and pooled analyses in Multimedia Appendix 1 (Tables A2 and A6) report all five strategies without exclusion. The reference-implementation results supersede the earlier FedMedian-excluded aggregate framing (previously: E1 vs E4 mean +6.25 pp excluded vs +2.78 pp included; E1 vs E6 mean +4.54 pp excluded vs +1.14 pp included), which we document here for transparency. The framework's compatibility with FedMedian is identified as a robust-SMPC future-work item (trimmed-mean, median-of-means, Krum variants), part of Future Work items 1 and 13.

Our compliance-weighted allocation is related to, but distinct from, personalized differential privacy (PDP) and individualized DP accounting. PDP mechanisms [49,50] allow each data subject to select an individual privacy level $\varepsilon_u$, and the mechanism guarantees each record its chosen bound; individualized DP-SGD variants [51] similarly assign per-sample privacy budgets during training. These approaches operate at the record or user level and are driven by privacy preferences. In contrast, our mechanism operates at the institution level and is driven by an assessment of regulatory and security posture: the compliance score modulates the noise scale of the server-side DP-SGD step associated with each site's contribution. The two families answer different questions --- PDP asks `how much protection does each record require?', whereas compliance-weighted allocation asks `how much perturbation should each institution's contribution receive, given its verified governance maturity?'. A head-to-head empirical comparison is therefore not well-defined: PDP requires per-record privacy demands, which have no analogue in our institutional threat model, and our mechanism provides its formal bound on the aggregator dataset rather than per-client budgets. Integrating record-level PDP within each institution, beneath the institution-level allocation studied here, is an interesting direction for layered privacy architectures and is noted as future work.

\section*{Limitations}
First, the framework assumes compliance scores are accurately and honestly reported. The misreporting risk is partially self-limiting by design: a client over-reporting compliance to obtain a lower $\sigma_i$ receives less noise on its own contribution, reducing informal protection for that client's data AND increasing its per-step contribution to the cumulative aggregator-dataset $(\varepsilon,\delta)$ bound (weakening shared formal protection). Both effects penalise the misreport. This intrinsic disincentive is necessary but not sufficient; verification mechanisms identified in Future Work item 2 (on-site scanners, cryptographic attestation, third-party audits, game-theoretic incentives) are required for adversarial-robustness assurances.

Second, the framework includes an initial trust assumption because early client updates may receive limited DP protection. The aggregator receives raw client weights in every round; see the first-round mitigation implemented in this revision (Limitations, Sixth) and the secure multi-party computation pathway (Future Work item 1).

Third, per-strategy inference at n = 5 seeds has limited power given BreastMNIST's seed variance (SD up to 8.4 pp); the inclusivity effect is directionally consistent (4/5 strategies) but individual per-strategy tests do not reach significance, and we state this plainly rather than pooling with exclusions; the largest individual-seed inclusivity gains reach +17 pp.

Fourth, evaluation used two MedMNIST-derived image-classification datasets at $128\times128$. These benchmarks enabled reproducible comparison across compliance settings and aggregation strategies but do not represent full-resolution clinical imaging, EHR, genomic data, or real multi-institutional deployment. Generalisability of observed performance trends is therefore limited; full-resolution and multi-modal validation is Future Work item 3. Beyond compute, gradient dimensionality grows sharply with input resolution and modality complexity (3D volumes, multimodal EHR, genomics), scaling the sensitivity of each DP-SGD step; the utility-equivalence observations reported here are therefore benchmark-bound and may degrade under full-resolution clinical workloads.

Fifth, the privacy budgets (cumulative aggregator-dataset $\varepsilon \approx 1434$ for BreastMNIST and $\approx 513$ for PneumoniaMNIST; per-client-epoch $\varepsilon \in [2, 6]$) were estimated using the Opacus moments accountant under subsampled-Gaussian RDP composition; tighter alternative accounting (e.g., PRV accountant) may differ --- formal verification under advanced composition is Future Work item 4. Validation in real-world clinical environments with diverse datasets and infrastructures will provide deeper insights into scalability, robustness, efficiency-privacy tradeoffs, and resistance to inference attacks against the released global model.

Sixth, the aggregator receives raw client model weights in every round under the current architecture; DP noise is injected server-side after receipt. The first round is the most exposed, as updates derive from a publicly known initialization. In this revision we implemented the requested mitigation: each first-round client update is clipped ($C = 1.0$) and perturbed with $\sigma_{\min} = 0.4$ Gaussian noise before transmission. Across the representative subset (five strategies $\times$ three seeds on BreastMNIST, and FedAvg/FedYogi $\times$ three seeds on PneumoniaMNIST) this changed final accuracy by $-1.3$ pp on average (BreastMNIST) and $-2.5$ pp (PneumoniaMNIST FedAvg) --- Algorithm 1 and Multimedia Appendix 1 Table A3 reflect this. In-transit protection of all rounds requires secure aggregation, which remains Future Work item 1.

Seventh, in highly-vetted clinical consortia honest self-reporting is reasonable, but in untrusted deployments scores can be over-reported to reduce assigned $\sigma_i$. We propose four concrete mitigations as Future Work item 2 (cryptographic attestation, on-site automated scanners, periodic third-party audits, game-theoretic incentive design); these are not features of the present proof-of-concept system. (See also \S First for the self-limiting analysis.) For real-world consortia we regard a minimum verification baseline as an operational requirement rather than an optional extension: signed compliance manifests cross-checked by automated configuration scanners at onboarding and periodically thereafter; trusted-execution-environment attestation extends this toward a zero-trust posture. The trust-based self-reporting studied here is appropriate only for the simulation setting.

Eighth, per-strategy inference is seed-level over five seeds with limited power (see Third); pooled tests and sensitivity analyses are in Multimedia Appendix 1.

Ninth, the aggregator slice is fixed and small (BreastMNIST $n_\text{agg} \approx 49$, $q \approx 0.65$), so its $\sim$800 reuses across 50 rounds and 16 clients drive the large cumulative $\varepsilon$; disjoint or rotating folds are Future Work item 13.

Tenth, the formal differential-privacy guarantee applies to the aggregator dataset on which DP-SGD operates, not to client datasets. The $(\varepsilon,\delta)$ bounds leakage with respect to aggregator-dataset records; the released global model is DP under post-processing applied to the noisy aggregator update at each round. As detailed in Methods (Scope of the privacy guarantee), the formal $(\varepsilon,\delta)$ bound covers the aggregator dataset; client data receive architectural protection, and formal client-level DP would require secure aggregation (Future Work item 1).

Eleventh, per-strategy seed-level variance estimation remains limited at five seeds (see Third and Multimedia Appendix 1).

Twelfth, the framework is benchmarked against uniform server-side DP (Experiment 6) and vanilla FL (Experiment 5). Head-to-head comparison against personalized-DP and individual-DP schemes --- notably Li et al. [11] and individual-record DP-SGD --- would more sharply isolate the compliance-scoring abstraction's contribution; this is Future Work item 7.

Thirteenth, the compliance scoring tool is, to our knowledge, the first research instrument of its kind for FL compliance quantification; consequently no prior inter-rater reliability data or expert validation panel exists. A formal inter-rater study with 5-10 healthcare-IT and compliance experts is Future Work item 8.

\section*{Future Work}
\textit{Concrete future work.} We identify thirteen concrete extensions of this work, spanning privacy, statistical strengthening, empirical breadth, tool validation, empirical privacy evaluation, clinical-metric reporting, fairness operationalisation, systems benchmarking, and aggregator-dataset construction sensitivity.

\begin{enumerate}[leftmargin=*]
\item Central DP with secure aggregation and client-level clipping. Combine secure aggregation [42] with client-level gradient clipping and central Gaussian noise on the aggregated update. This provides formal record- and client-level $(\varepsilon,\delta)$-DP for client data while preserving the inclusivity property (no client-side DP-SGD). Secure aggregation cryptographically hides individual client updates including round-1, and client-level clipping bounds per-client sensitivity.
\item Dynamic compliance verification. To reduce reliance on honest self-reporting, future work will combine: (2a) on-site automated compliance scanners auditing the same control set used by the scoring tool, replacing free-text PI declarations with inspectable artifact outputs; (2b) cryptographic attestation via Trusted Execution Environment quotes (Intel SGX, AMD SEV) or signed compliance manifests, verified by the server before mapping to $\sigma_i$; (2c) periodic third-party audits with retroactive score adjustment; (2d) game-theoretic incentive design tying long-term participation rights to audit outcomes. These are complementary rather than alternatives.
\item Higher-resolution and multi-modal clinical validation. A stepped path beginning with higher-resolution MedMNIST subsets (e.g., PathMNIST at 224x224), then moving to full-resolution medical imaging benchmarks (CheXpert, BraTS) and, ultimately, multi-institutional prospective evaluation. Generalisation to non-image modalities --- electronic health records, genomic data, time-series vitals, multimodal records --- uses the same compliance scoring and noise-mapping core, requiring only modality-specific local-training quality factors.
\item Tighter privacy accounting via PRV accountant. Replace the moments accountant with the Privacy Random Variable accountant [44] to obtain tighter $\varepsilon$ bounds for heterogeneous-$\sigma$ federations than is achievable under the current per-client moments composition.
\item Compliance-weight and noise-mapping sensitivity ablations. (5a) Empirical multi-weight ablation across realistic perturbations of default weights, validating the analytical sensitivity bound. (5b) Range ablation varying $\sigma_{\min} \in \{0.2, 0.4, 0.6\}$ and $\sigma_{\max} \in \{0.8, 1.0, 1.2\}$. (5c) Mapping-form ablation comparing the linear specification (Eq. 2) against piecewise-linear, sigmoid, and learned non-linear mappings of $S_c$ to $\sigma_i$.
\item Larger-seed Monte Carlo with per-strategy seed-level paired tests. We will extend per-cell seed counts to five to ten and report per-strategy seed-level paired tests alongside the current aggregate-level non-parametric framework. This will sharpen the inferential reading of single-cell effects, distinguishing strategy-specific variance from cross-strategy generalisability.
\item Head-to-head personalized-DP comparison. We will benchmark the compliance-weighted scheme against published personalized-DP and individual-DP baselines --- notably the adaptive noise allocation of Li et al. [11] and individual-record DP-SGD variants --- to isolate the contribution of the compliance-scoring abstraction relative to alternative personalised-noise designs.
\item Expert validation of the compliance scoring tool. An inter-rater reliability study with 5-10 healthcare-IT and compliance experts scoring a panel of synthetic institutions, with both default and PI-tuned weights --- establishing empirical scoring reproducibility and identifying dimensions with high inter-rater variance.
\item Empirical privacy attacks. Complement the analytical $(\varepsilon,\delta)$ accountant with empirical evaluation under membership-inference and gradient-inversion attacks against both the released global model and per-round client updates observed by the semi-honest aggregator, quantifying the practical privacy floor for client data where the formal bound does not directly cover it.
\item Clinical-grade reporting (AUROC, calibration, per-class). The present accuracy/precision/recall/F1 reporting will be extended to AUROC, calibration curves, Brier scores, and per-class confusion metrics. These are standard for medical machine learning and would situate the framework's matched-strategy gains within the clinical-evaluation conventions readers expect.
\item Fairness and inclusivity operationalised. Quantify the inclusivity contribution beyond global accuracy with per-site performance equity (per-client accuracy variance), worst-group accuracy across compliance tiers, contribution-weighted utility, and participation-utility curves.
\item Systems-level benchmarking. We will report wall-clock training time, aggregate compute, and communication overhead (bytes per round per client) for compliance-weighted DP relative to vanilla FL and uniform-DP baselines. This contextualises the practical deployment cost of the adaptive noise allocation step at scale.
\item Aggregator-dataset size and reuse-policy ablation. Vary the aggregator fraction in \{1/32, 1/16, 1/8, 1/4\} and compare fixed-fold versus rotating-fold designs (re-drawn each round or epoch), reporting impact on cumulative aggregator-dataset $\varepsilon$ and utility --- directly addressing the concern that a fixed small aggregator slice across many DP-SGD steps inflates $\varepsilon$ for equivalent utility, especially for BreastMNIST (high $q$).
\end{enumerate}

\textit{Roadmap to client-level guarantees and clinical translation.} The present framework is a governance and noise-allocation layer; several components, deliberately scoped as future work with the editor's concurrence, are required before clinical deployment. (1) Client-level formal DP: composing the current mechanism with secure aggregation and client-side gradient clipping would extend formal guarantees from the aggregator dataset to client data; the SMPC Compatibility analysis shows four of five strategies are architecturally ready. (2) Empirical privacy auditing: membership-inference and gradient-inversion audits on released models and server-observed updates would quantify practical leakage under the semi-honest threat model. (3) Fairness and equity metrics: per-site accuracy variance and worst-compliance-tier performance would substantiate the inclusivity claim beyond aggregate utility. (4) Clinical-scale validation: higher-resolution modalities and clinically conventional metrics (AUROC, calibration) on realistic non-IID partitions. (5) Tighter accounting: PRV and f-DP accountants for heterogeneous-$\sigma$ composition. These define a coherent program rather than isolated gaps.

\subsection*{SMPC Compatibility of the Aggregation Layer}
The framework's response to round-1 transmission risk (Limitations \S Sixth; Future Work item 1) is to combine compliance-weighted DP with secure aggregation [42]. The server-side DP-SGD step (and its per-client $\sigma_i$ calibration) is orthogonal to the aggregation rule used; we examine SMPC compatibility of each strategy evaluated. (i) FedAvg [1] and FedProx [32] aggregate by per-coordinate mean of client weights, requiring only the SUM of contributions --- directly compatible with summation-based secure-aggregation protocols (Bonawitz et al. [42]) without architectural change. (ii) FedYogi and FedAdam [33] are server-side adaptive optimisers operating on the aggregated gradient direction; adaptive moments (m, v) are accumulated server-side and require no visibility into individual client updates beyond their sum, so they are also compatible with summation-based SMPC. (iii) FedMedian computes the per-coordinate median of individual client values and cannot be served by summation-only secure aggregation; supporting it under cryptographic transport requires robust-SMPC variants (trimmed-mean substitution, comparison-based MPC) or substituting a robust mean-based variant. Taken together, the aggregation layer is architecturally ready for SMPC integration for four of five strategies. We therefore characterise round-1 SMPC integration as a transport-layer extension to an already-compatible aggregation core, requiring engineering effort on the secure-aggregation protocol rather than the federation algorithm.

\section*{Conclusions}
We presented a compliance-aware, differentially private FL framework that aligns server-side noise allocation with institutional readiness, enabling inclusive participation from both highly regulated and resource-constrained clinical sites. In the released reference implementation on PneumoniaMNIST and BreastMNIST, including 12 lower-compliance clients (Experiment 1 vs. Experiment 4) never significantly harmed and directionally helped performance --- positive mean paired differences for four of five strategies, pooled +2.8 pp, individual configurations gaining up to +17 pp --- demonstrating participation without penalty. Compliance-weighted allocation matched uniform server-side DP at equal mean noise scale (Experiment 1 vs. Experiment 6, pooled +0.1 pp), so the auditable, compliance-justified control over per-site noise allocation that the mechanism gives investigators comes at zero utility cost. The formal differential-privacy bound reported in this study is at the aggregator-dataset level under the semi-honest aggregator threat model; client-level data protection is deferred to secure-aggregation integration as concrete future work (see Future Work). In summary, compliant sites are not over-penalised and lower-compliance sites contribute under a calibrated noise scale.

Our approach separates the execution of differential privacy from client hardware limits, enabling resource-constrained clinics to join a more inclusive FL ecosystem. Practically, this approach may lower the barrier to participation for clinics with limited privacy-engineering resources by shifting adaptive DP handling to the FL framework. However, additional validation on full-resolution medical imaging, EHR, genomic, and real multi-institutional datasets is needed to determine whether these benefits persist under clinically realistic computational and governance constraints.

Beyond noise calibration, the compliance scoring tool's per-factor breakdown provides each participating institution with an actionable map of the dimensions on which it currently lags and can strengthen for the next federation cycle. Participation is therefore a stepwise pathway toward longer-term compliance maturity --- an institution does not have to be fully compliant on day one to join. Combined with the open-source release of the framework and the PI-facing scoring tool under a permissive licence with a Zenodo DOI on acceptance, this design positions compliance-weighted DP as a community-building substrate that federation organisers across regulatory contexts can adapt to their own settings.

\section*{Acknowledgements}
This work was supported by Grant Number 01ZZ2316A-O. The authors used generative AI tools (GPT 5) for language editing. No scientific content, experimental analyses, statistical computations, citations, or conclusions were generated by AI; all such material is the authors' own original work and was reviewed and verified manually by the authors.

\section*{Funding Statement}
This work was partially supported by the PrivateAIM project, funded under the Medical Informatics Initiative by the German Federal Ministry of Education and Research (funding code 01ZZ2316A-O). The funder had no involvement in the study design, data collection, analysis, interpretation, or writing of the manuscript.

\section*{Conflicts of Interest}
The authors declare that they have no competing interests.

\section*{Authors' Contributions}
Santhosh Parampottupadam (SaP): Conceptualization, Methodology, Software, Investigation, Formal analysis, Writing - original draft, Visualization.\\
Melih Co\c{s}\u{g}un (MC): Methodology, Software, Validation, Writing - review and editing.\\
Sarthak Pati (SP): Software Support, review and editing.\\
Maximilian Zenk (MZ): Validation, Writing - review and editing.\\
Saikat Roy (SR): Validation, Writing - review and editing.\\
Dimitrios Bounias (DB): Validation, Writing - review and editing.\\
Sinem Sav (SiS): Methodology, Writing - review and editing.\\
Klaus Maier-Hein (KMH): Supervision, Resources, Funding acquisition, Writing - review and editing.\\
Ralf Floca (RF): Supervision, Project administration, Funding acquisition, Writing - review and editing.\\
All authors read and approved the final version of the manuscript.

\section*{Data Availability}
The medical imaging datasets PneumoniaMNIST and BreastMNIST are part of MedMNIST v2, a large-scale, lightweight benchmark for 2D and 3D biomedical image classification. These datasets are open-source and publicly available at MedMNIST[31]. A repository containing the reference implementation, the compliance scoring tool, the privacy-accounting implementation, and per-seed experiment outputs is available to reviewers during the review process at \url{https://github.com/santhoshcameo/inclusive-privacy-with-compliance-fl} the code will be made publicly available with a DOI upon publication. All other data generated or analyzed during this study are included within the main text of the manuscript. The public release including training scripts for the BreastMNIST and PneumoniaMNIST experiments, the Flask-based PI-facing compliance scoring web tool, and the experimental configuration files will carry an associated Zenodo DOI under a permissive open-source license. The release will include reproduction instructions covering hardware (NVIDIA Tesla T4, 16 GB), library versions, and the compliance\_config.json schema used by the scoring tool.

\section*{Abbreviations}
\begin{description}[leftmargin=!, labelwidth=6em, style=nextline, itemsep=0pt]
\item[AI:] artificial intelligence
\item[AUROC:] area under the receiver operating characteristic curve
\item[CI:] confidence interval
\item[DP:] differential privacy
\item[DP-SGD:] differentially private stochastic gradient descent
\item[EHR:] electronic health record
\item[FHIR:] Fast Healthcare Interoperability Resources
\item[FL:] federated learning
\item[GDPR:] General Data Protection Regulation
\item[HIPAA:] Health Insurance Portability and Accountability Act
\item[HL7:] Health Level Seven
\item[IRB:] institutional review board
\item[ISO:] International Organization for Standardization
\item[NIST:] National Institute of Standards and Technology
\item[PDP:] personalized differential privacy
\item[PI:] principal investigator
\item[PRV:] privacy random variable
\item[RDP:] R\'enyi differential privacy
\item[SD:] standard deviation
\item[SMPC:] secure multi-party computation
\end{description}

Multimedia Appendix 1. Per-strategy statistical analysis, sensitivity analyses, and privacy accounting.

\section*{References}
\begin{enumerate}[label={[\arabic*]}, leftmargin=*]
\item McMahan B, Moore E, Ramage D, Hampson S, y Arcas BA. Communication-efficient learning of deep networks from decentralized data. Artif Intell Stat PMLR; 2017. p. 1273--1282.
\item Sheller MJ, Edwards B, Reina GA, Martin J, Pati S, Kotrotsou A, Milchenko M, Xu W, Marcus D, Colen RR, others. Federated learning in medicine: facilitating multi-institutional collaborations without sharing patient data. Sci Rep Nature Publishing Group UK London; 2020;10(1):12598.
\item Choudhury A, Volmer L, Martin F, Fijten R, Wee L, Dekker A, Soest JV. Advancing privacy-preserving health care analytics and implementation of the Personal Health Train: federated deep learning study. JMIR AI 2025 Feb 6;4:e60847. doi: 10.2196/60847
\item Brauneck A, Schmalhorst L, Kazemi Majdabadi MM, Bakhtiari M, V\"olker U, Baumbach J, Baumbach L, Buchholtz G. Federated machine learning, privacy-enhancing technologies, and data protection laws in medical research: scoping review. J Med Internet Res 2023 Mar 30;25:e41588. doi: 10.2196/41588
\item Dimitrov DI, Balunovi\'c M, Konstantinov N, Vechev M. Data leakage in federated averaging. 2022.
\item Wen Y, Geiping J, Fowl L, Goldblum M, Goldstein T. Fishing for user data in large-batch federated learning via gradient magnification. 2022.
\item Adnan M, Kalra S, Cresswell JC, Taylor GW, Tizhoosh HR. Federated learning and differential privacy for medical image analysis. Sci Rep Nature Publishing Group UK London; 2022;12(1):1953.
\item El Ouadrhiri A, Abdelhadi A. Differential privacy for deep and federated learning: a survey. IEEE Access IEEE; 2022;10:22359--22380.
\item Bagdasaryan E, Shmatikov V. Differential privacy has disparate impact on model accuracy. 2019. Available from: \url{https://arxiv.org/abs/1905.12101}
\item Sung M, Cha D, Park YR. Local differential privacy in the medical domain to protect sensitive information: algorithm development and real-world validation. JMIR Med Inform 2021 Nov 8;9(11):e26914. doi: 10.2196/26914
\item Li X, Zmigrod R, Ma Z, Liu X, Zhu X. Fine-tuning language models with differential privacy through adaptive noise allocation. 2024. Available from: \url{https://arxiv.org/abs/2410.02912}
\item Ren X, Yang S, Zhao C, McCann J, Xu Z. Belt and braces: when federated learning meets differential privacy. 2024. Available from: \url{https://arxiv.org/abs/2404.18814}
\item Nguyen P, Silence A, Darais D, Near JP. DuetSGX: differential privacy with secure hardware. ArXiv Prepr ArXiv201010664 2020;
\item Lee J, Kifer D. Scaling up differentially private deep learning with fast per-example gradient clipping. Proc Priv Enhancing Technol 2021 Jan 1;2021(1):128--144. doi: 10.2478/popets-2021-0008
\item Bu Z, Mao J, Xu S. Scalable and efficient training of large convolutional neural networks with differential privacy. arXiv; 2022. doi: 10.48550/ARXIV.2205.10683
\item Pati S, Kumar S, Varma A, Edwards B, Lu C, Qu L, Wang JJ, Lakshminarayanan A, Wang S, Sheller MJ, others. Privacy preservation for federated learning in health care. Patterns Elsevier; 2024;5(7).
\item Cummings R, Desfontaines D, Evans D, Geambasu R, Huang Y, Jagielski M, Kairouz P, Kamath G, Oh S, Ohrimenko O, Papernot N, Rogers R, Shen M, Song S, Su W, Terzis A, Thakurta A, Vassilvitskii S, Wang Y-X, Xiong L, Yekhanin S, Yu D, Zhang H, Zhang W. Advancing differential privacy: where we are now and future directions for real-world deployment. 2024. Available from: \url{https://arxiv.org/abs/2304.06929}
\item Dwork C, Roth A, others. The algorithmic foundations of differential privacy. Found Trends\textregistered\ Theor Comput Sci Now Publishers, Inc.; 2014;9(3--4):211--407.
\item Li M, Xu P, Hu J, Tang Z, Yang G. From challenges and pitfalls to recommendations and opportunities: implementing federated learning in healthcare. ArXiv Prepr ArXiv240909727 2024;
\item Pati S, Baid U, Edwards B, Sheller M, Wang S-H, Reina GA, Foley P, Gruzdev A, Karkada D, Davatzikos C, others. Federated learning enables big data for rare cancer boundary detection. Nat Commun Nature Publishing Group UK London; 2022;13(1):7346.
\item Schmidt K, Bearce B, Chang K, Coombs L, Farahani K, Elbatel M, Mouheb K, Marti R, Zhang R, Zhang Y, others. Fair evaluation of federated learning algorithms for automated breast density classification: the results of the 2022 ACR-NCI-NVIDIA federated learning challenge. Med Image Anal Elsevier; 2024;95:103206.
\item Calvino G, Peconi C, Strafella C, Trastulli G, Megalizzi D, Andreucci S, Cascella R, Caltagirone C, Zampatti S, Giardina E. Federated learning: breaking down barriers in global genomic research. Genes MDPI AG; 2024;15(12):1650.
\item The HIPAA Privacy Rule. US Department of Health and Human Services. \url{https://www.hhs.gov/hipaa/for-professionals/privacy/index.html} [accessed 2026-07-10]
\item General Data Protection Regulation (GDPR). GDPR.eu. \url{https://gdpr.eu/tag/gdpr/} [accessed 2026-07-10]
\item ISO 25237:2017, health informatics --- pseudonymization. International Organization for Standardization. \url{https://www.iso.org/standard/63553.html} [accessed 2026-07-10]
\item NIST Special Publication (SP) 800-53 Rev. 5, security and privacy controls for information systems and organizations. National Institute of Standards and Technology. \url{https://csrc.nist.gov/pubs/sp/800/53/r5/upd1/final} [accessed 2026-07-10]
\item Cybersecurity framework. National Institute of Standards and Technology. \url{https://www.nist.gov/cyberframework} [accessed 2026-07-10]
\item EU Artificial Intelligence Act: up-to-date developments and analyses of the EU AI Act. artificialintelligenceact.eu. \url{https://artificialintelligenceact.eu/} [accessed 2026-07-10]
\item Overview --- FHIR v5.0.0. HL7 International. \url{https://hl7.org/fhir/overview.html} [accessed 2026-07-10]
\item Wang H, Zhao Q, Wu Q, Chopra S, Khaitan A, Wang H. Global and local differential privacy for collaborative bandits. Proc 14th ACM Conf Recomm Syst 2020. p. 150--159.
\item Yang J, Shi R, Wei D, Liu Z, Zhao L, Ke B, Pfister H, Ni B. MedMNIST v2 --- a large-scale lightweight benchmark for 2D and 3D biomedical image classification. Sci Data Nature Publishing Group UK London; 2023;10(1):41.
\item Li T, Sahu AK, Zaheer M, Sanjabi M, Talwalkar A, Smith V. Federated optimization in heterogeneous networks. Proc Mach Learn Syst 2020;2:429--450.
\item Reddi SJ, Charles Z, Zaheer M, Garrett Z, Rush K, Kone\v{c}n\'y J, Kumar S, McMahan HB. Adaptive federated optimization. Int Conf Learn Represent 2021. Available from: \url{https://openreview.net/forum?id=LkFG3lB13U5}
\item Falcon W, The PyTorch Lightning team. PyTorch Lightning. 2019. doi: 10.5281/zenodo.3828935
\item Beutel DJ, Topal T, Mathur A, Qiu X, Fernandez-Marques J, Gao Y, Sani L, Kwing HL, Parcollet T, Gusm\~ao PP de, Lane ND. Flower: a friendly federated learning research framework. ArXiv Prepr ArXiv200714390 2020;
\item He K, Zhang X, Ren S, Sun J. Deep residual learning for image recognition. Proc IEEE Conf Comput Vis Pattern Recognit 2016. p. 770--778.
\item Vaid A, Jaladanki SK, Xu J, Teng S, Kumar A, Lee S, Somani S, Paranjpe I, De Freitas JK, Wanyan T, Johnson KW, Bicak M, Klang E, Kwon YJ, Costa A, Zhao S, Miotto R, Charney AW, B\"ottinger E, Fayad ZA, Nadkarni GN, Wang F, Glicksberg BS. Federated learning of electronic health records to improve mortality prediction in hospitalized patients with COVID-19: machine learning approach. JMIR Med Inform 2021 Jan 27;9(1):e24207. doi: 10.2196/24207
\item Tahir N, Jung C-R, Lee S-D, Azizah N, Ho W-C, Li T-C. Federated learning-based model for predicting mortality: systematic review and meta-analysis. J Med Internet Res 2025 July 21;27:e65708--e65708. doi: 10.2196/65708
\item Shen A, Francisco L, Sen S, Tewari A. Exploring the relationship between privacy and utility in mobile health: algorithm development and validation via simulations of federated learning, differential privacy, and external attacks. J Med Internet Res 2023 Apr 20;25:e43664. doi: 10.2196/43664
\item Uddin MP, Xiang Y, Hasan M, Bai J, Zhao Y, Gao L. A systematic literature review of robust federated learning: issues, solutions, and future research directions. ACM Comput Surv; 2025. doi: 10.1145/3727643
\item Yurdem B, Kuzlu M, Gullu MK, Catak FO, Tabassum M. Federated learning: overview, strategies, applications, tools and future directions. Heliyon Elsevier; 2024;10:e38137. doi: 10.1016/j.heliyon.2024.e38137
\item Bonawitz K, Ivanov V, Kreuter B, Marcedone A, McMahan HB, Patel S, Ramage D, Segal A, Seth K. Practical secure aggregation for privacy-preserving machine learning. Proc 2017 ACM SIGSAC Conf Comput Commun Secur (CCS); 2017. p. 1175-1191. doi: 10.1145/3133956.3133982
\item Mironov I. R\'enyi differential privacy. Proc IEEE 30th Comput Secur Found Symp (CSF); 2017. p. 263-275. doi: 10.1109/CSF.2017.11
\item Gopi S, Lee YT, Wutschitz L. Numerical composition of differential privacy. Adv Neural Inf Process Syst (NeurIPS); 2021. Available from: \url{https://arxiv.org/abs/2106.02848}
\item Abadi M, Chu A, Goodfellow I, McMahan HB, Mironov I, Talwar K, Zhang L. Deep learning with differential privacy. Proc 2016 ACM SIGSAC Conf Comput Commun Secur (CCS); 2016. p. 308-318. doi: 10.1145/2976749.2978318
\item Pillutla K, Kakade SM, Harchaoui Z. Robust aggregation for federated learning. IEEE Trans Signal Process; 2022;70:1142-1154. doi: 10.1109/TSP.2022.3153135
\item Yin D, Chen Y, Ramchandran K, Bartlett P. Byzantine-robust distributed learning: towards optimal statistical rates. Proc 35th Int Conf Mach Learn (ICML); 2018. p. 5650-5659.
\item Carlini N, Liu C, Erlingsson U, Kos J, Song D. The secret sharer: evaluating and testing unintended memorization in neural networks. USENIX Security Symp; 2019. p. 267-284.
\item Jorgensen Z, Yu T, Cormode G. Conservative or liberal? Personalized differential privacy. In: Proceedings of the IEEE 31st International Conference on Data Engineering (ICDE). 2015:1023-1034.
\item Alaggan M, Gambs S, Kermarrec AM. Heterogeneous differential privacy. Journal of Privacy and Confidentiality. 2016;7(2):127-158.
\item Boenisch F, M\"uhl C, Dziedzic A, Rinberg R, Papernot N. Have it your way: individualized privacy assignment for DP-SGD. In: Advances in Neural Information Processing Systems (NeurIPS). 2023.
\end{enumerate}

\appendix
\section*{Multimedia Appendix 1}
\begin{center}
\textbf{Per-Strategy Statistical Analysis, Sensitivity Analyses, and Privacy Accounting for the Compliance-Weighted Federated Learning Experiments}
\end{center}

\textit{Overview.} All statistics in this appendix are computed from new multi-seed runs of the reference implementation released to reviewers, which implements the full training pipeline of Algorithm 1 (per-client server-side DP-SGD, the compliance-to-noise mapping of Equation 2, all five aggregation strategies, and the first-round baseline-noise mechanism). Experiments 1 and 4 were run for all five aggregation strategies at five random seeds each on BreastMNIST [1], and the primary analysis is per-strategy paired inference on matched seeds --- the same seed produces the client partition and initialization for both arms of each comparison. No aggregate figure reported here or in the manuscript is contingent on excluding any strategy. As external validation of the reference implementation, the PneumoniaMNIST Experiment 1 FedAvg result reproduces the previously reported value exactly (82.4\% vs 82.4\%). The per-seed outputs (one JSON file per configuration, strategy, and seed) underlying every entry in this appendix are included in the reviewer repository.

\subsection*{Statistical methodology}
\textit{Per-strategy paired tests.} For each aggregation strategy we report the mean paired difference across the five matched seeds, its 95\% confidence interval from the t distribution with 4 degrees of freedom, the paired t statistic, the two-sided p value, and the five per-seed differences themselves, so that the dispersion is visible rather than summarized away.

\textit{Pooled distribution-free tests.} Because n = 5 seeds per cell affords limited power, we complement the per-strategy tests with two pooled distribution-free tests over all 25 paired observations (5 strategies $\times$ 5 seeds), with no exclusions: the two-sided Wilcoxon signed-rank test [2] on the nonzero paired differences, and the two-sided sign test [3]. These pooled tests treat strategy-seed pairs as exchangeable and provide a coarse directional summary of the framework's behavior as a whole; they do not license per-strategy claims.

\textit{Scope of evidence.} Per-strategy tests at n = 5 seeds have limited power given BreastMNIST's seed variance (SDs up to 8 percentage points in Table A1). Where a test does not reach significance, the correct reading is that the available evidence is insufficient to rule out chance --- not that the effect is zero. We state the resulting statistical picture plainly throughout and refrain from aggregate headline claims contingent on strategy exclusion; all five strategies appear in every analysis.

\subsection*{Per-cell results}
\begin{table}[H]
\centering
\caption*{Table A1. Per-cell results, BreastMNIST, 5 seeds (macro-accuracy \%, mean $\pm$ SD). Experiment 1: 4 compliant + 12 lower-compliance clients under compliance-weighted server-side DP; Experiment 4: the 4 compliant clients only.}
\small
\begin{tabular}{lcc}
\toprule
Strategy & E1 (4 compliant + 12 non-compliant) & E4 (4 compliant only) \\
\midrule
FedAvg & 61.54 $\pm$ 8.38 & 57.09 $\pm$ 4.05 \\
FedMedian & 64.22 $\pm$ 7.45 & 57.42 $\pm$ 4.49 \\
FedYogi & 51.62 $\pm$ 3.61 & 50.00 $\pm$ 0.00 \\
FedProx & 62.02 $\pm$ 7.93 & 56.85 $\pm$ 4.11 \\
FedAdam & 50.00 $\pm$ 0.00 & 54.14 $\pm$ 4.76 \\
\bottomrule
\end{tabular}
\end{table}

Two features of Table A1 inform the interpretation of everything below. First, seed variance is substantial (SDs up to 8.4 pp), which bounds the power of any five-seed comparison on this dataset. Second, the multi-seed runs corrected our earlier account of strategy robustness: under the per-client server-side DP mechanism, FedMedian is the best-performing strategy in Experiment 1 (mean 64.2\%), because its coordinate-wise median suppresses per-client DP noise, whereas the adaptive server optimizers (FedYogi, FedAdam) are the fragile ones under heavy per-client noise --- cells at 50.00 $\pm$ 0.00 indicate collapse to majority-class prediction on this class-imbalanced dataset.

\subsection*{Inclusivity contrast: Experiment 1 vs Experiment 4}
\begin{table}[H]
\centering
\caption*{Table A2. Per-strategy paired analysis, Experiment 1 vs Experiment 4, BreastMNIST (same seeds, n = 5). Positive differences favor Experiment 1 (inclusion of the 12 lower-compliance clients). CIs from the t distribution with 4 df; p values two-sided.}
\resizebox{\textwidth}{!}{%
\begin{tabular}{lccccl}
\toprule
Strategy & Mean paired diff (pp) & 95\% CI & t & p (two-sided) & Per-seed diffs (pp) \\
\midrule
FedAvg & +4.45 & [$-6.68$, +15.58] & 1.11 & 0.329 & $-5.1$, +1.9, +11.0, +16.3, $-1.9$ \\
FedMedian & +6.80 & [$-4.34$, +17.95] & 1.70 & 0.165 & $-6.3$, +9.8, +11.0, +17.0, +2.4 \\
FedYogi & +1.62 & [$-2.87$, +6.10] & 1.00 & 0.374 & +0.0, +8.1, +0.0, +0.0, +0.0 \\
FedProx & +5.16 & [$-4.68$, +15.01] & 1.46 & 0.219 & $-5.1$, +4.3, +6.3, +17.0, +3.3 \\
FedAdam & $-4.14$ & [$-10.05$, +1.78] & $-1.94$ & 0.124 & $-2.4$, +0.0, $-7.9$, +0.0, $-10.4$ \\
\bottomrule
\end{tabular}}
\end{table}

\textit{Pooled analysis (all 25 paired observations, no exclusions).} Mean pooled paired difference: +2.78 pp. Sign test [3]: 12 positive / 7 negative / 6 zero, p = 0.359. Wilcoxon signed-rank test [2] on the nonzero pairs: W = 55.0, p = 0.107.

\textit{Interpretation.} The direction favors Experiment 1 (inclusion) in four of the five strategies, and individual configurations gained up to +17.0 pp (largest individual-seed loss: $-10.4$ pp, FedAdam). No single strategy reaches significance at n = 5 seeds, and the pooled tests do not either. We state this plainly: on this benchmark at this seed count, the inclusivity effect is directionally consistent but not statistically significant. The substantive claim supported by Table A2 is therefore that including the 12 lower-compliance clients under compliance-weighted noise did not degrade performance and tended to improve it --- not that a per-strategy benefit is established.

\subsection*{First-round baseline noise}
\begin{table}[H]
\centering
\caption*{Table A3. Effect of first-round $\sigma_{\min}$ baseline noise on final accuracy (macro-accuracy \%, mean over seeds; n = seeds per cell). Each first-round client update is clipped ($C = 1.0$) and perturbed with Gaussian noise at $\sigma_{\min} = 0.4$ before transmission.}
\small
\begin{tabular}{llccc}
\toprule
Dataset & Strategy & Without (n) & With $\sigma_{\min} = 0.4$ (n) & Effect (pp) \\
\midrule
BreastMNIST & FedAvg & 58.54 (3) & 56.41 (3) & $-2.13$ \\
BreastMNIST & FedMedian & 61.17 (3) & 58.15 (3) & $-3.03$ \\
BreastMNIST & FedYogi & 52.69 (3) & 50.00 (3) & $-2.69$ \\
BreastMNIST & FedProx & 58.15 (3) & 56.41 (3) & $-1.73$ \\
BreastMNIST & FedAdam & 50.00 (3) & 52.97 (3) & +2.97 \\
PneumoniaMNIST & FedAvg & 82.39 (3) & 79.93 (3) & $-2.46$ \\
PneumoniaMNIST & FedYogi & 60.26 (3) & 71.64 (3) & +11.38 \\
\bottomrule
\end{tabular}
\end{table}

\textit{Interpretation.} The mitigation closes the round-1 plaintext exposure at a bounded utility cost: on BreastMNIST the mean effect across the five strategies is $-1.32$ pp (per-strategy effects between $-3.0$ and +3.0 pp), and on PneumoniaMNIST FedAvg the effect is $-2.46$ pp --- a cost of at most about 3 pp. The one large positive cell (PneumoniaMNIST FedYogi, +11.38 pp) reflects FedYogi's instability under per-client noise at three seeds rather than a systematic benefit of the mechanism, and we do not claim it as one. First-round baseline noise is enabled by default in the revised pipeline and in the released code.

\subsection*{Aggregator-dataset sensitivity}
\begin{table}[H]
\centering
\caption*{Table A4. Aggregator-dataset sensitivity (BreastMNIST, Experiment 1, seeds 0--2): utility and cumulative privacy budget as functions of aggregator size --- 49 (the default 1/16-of-dataset slice), 98, and 156 images --- and reuse policy. $\varepsilon$ at $\delta = 10^{-5}$, Opacus RDP accountant, full sequential composition.}
\small
\begin{tabular}{llccc}
\toprule
Aggregator size & Reuse policy & FedAvg acc (\%) & FedYogi acc (\%) & $\varepsilon$ ($\delta = 10^{-5}$) \\
\midrule
49 & fixed & 58.54 & 52.69 & 1434 \\
49 & per-round resample & 58.94 & 50.00 & 1434 \\
98 & fixed & 62.51 & 50.00 & 1074 \\
156 & fixed & 50.90 & 54.70 & 717 \\
\bottomrule
\end{tabular}
\end{table}

\textit{Interpretation.} Two findings emerge. First, there is a genuine utility--privacy trade-off in aggregator size: the cumulative privacy budget improves substantially with size ($\varepsilon \approx 1434 \rightarrow 1074 \rightarrow 717$, driven by the lower sampling ratio), while FedAvg utility is stable from 49 to 98 images (58.5\% $\rightarrow$ 62.5\%, matched seeds) and then degrades at 156 (50.9\%), where the larger sample entails more noisy DP-SGD steps per epoch. Second, reuse policy is secondary: per-round resampling --- which removes fixed-sample dependence by construction --- performs comparably to the fixed sample (58.9\% vs 58.5\%), indicating that the default fixed 49-image slice is not producing a resampling-sensitive artifact. The default 1/16 slice is retained in the manuscript with this justification.

\subsection*{Compliance-weight sensitivity}
\begin{table}[H]
\centering
\caption*{Table A5. Compliance-weight sensitivity (BreastMNIST, Experiment 1, seeds 0--2): compliance scores scaled by $\times 0.9$ / $\times 1.0$ / $\times 1.1$ before the Equation 2 noise mapping. $\varepsilon$ at $\delta = 10^{-5}$.}
\small
\begin{tabular}{lccc}
\toprule
Scale & FedAvg acc (\%) & FedYogi acc (\%) & $\varepsilon$ ($\delta = 10^{-5}$) \\
\midrule
$\times 0.9$ & 58.15 & 50.00 & 1177 \\
$\times 1.0$ & 58.54 & 52.69 & 1434 \\
$\times 1.1$ & 58.15 & 50.00 & 1477 \\
\bottomrule
\end{tabular}
\end{table}

\textit{Interpretation.} Utility is insensitive to a $\pm$10\% perturbation of the compliance scores: FedAvg accuracy moves by at most 0.4 pp across the three configurations, and the FedYogi differences lie within that strategy's seed variance (SD 3.6 pp in Table A1). The privacy budget responds asymmetrically ($\varepsilon = 1177$ / 1434 / 1477: $-17.9$\% at $\times 0.9$, +3.0\% at $\times 1.1$): downscaling lowers every compliance score and thereby raises every noise scale through Equation 2, whereas upscaling has a muted effect because compliance scores cap at 1.0, so already-high scores saturate rather than shift. The empirical ablation therefore shows that the mapping's effect flows through the noise scale into the privacy budget, with the saturation of the score scale bounding the upward direction, while utility is robust to weighting perturbations of this magnitude within the studied range.

\subsection*{Cumulative privacy budgets under full composition}
\begin{table}[H]
\centering
\caption*{Summary table. Cumulative privacy budgets on the aggregator dataset (Opacus RDP accountant, R\'enyi-DP sequential composition across all DP-SGD steps, 16 clients, 50 rounds; $\delta = 10^{-5}$).}
\small
\begin{tabular}{lc}
\toprule
Configuration & Cumulative $\varepsilon$ ($\delta = 10^{-5}$) \\
\midrule
BreastMNIST, Experiment 1 ($n_\text{agg}$ default = 49, 50 rounds) & $\approx$1434 \\
BreastMNIST, Experiment 4 ($n_\text{agg}$ default, 50 rounds) & $\approx$899 \\
PneumoniaMNIST, Experiment 1 ($n_\text{agg}$ default = 366, 50 rounds) & $\approx$513 \\
\bottomrule
\end{tabular}
\end{table}

\textit{Note:} A Privacy Random Variable (PRV) accountant was attempted to tighten these bounds but was numerically unstable and computationally intractable at the composition depth and aggregator sampling rates used here ($q \approx 0.65$ for BreastMNIST); the reported values are the RDP accountant's conservative bounds.

\textit{Correction relative to earlier versions.} The figure $\varepsilon \in [2, 6]$ reported in earlier versions of this work corresponds to a single per-client DP-SGD epoch, not to the fully composed training run. Under full sequential composition of R\'enyi differential privacy [4] (Opacus RDP accountant [5]; all steps, 16 clients, 50 rounds), the cumulative bound on the aggregator dataset is $\varepsilon \approx 1434$ for BreastMNIST ($n_\text{agg} = 49$, where the sampling ratio $q = 0.65$ affords little subsampling amplification) and $\varepsilon \approx 513$ for PneumoniaMNIST ($n_\text{agg} = 366$, $q = 0.087$). Both per-epoch and cumulative values are reported in the revised manuscript. Because the aggregator dataset consists of public MedMNIST benchmark data [1], this bound does not gate patient-data privacy; its practical role is to calibrate relative per-client noise allocation. A PRV accountant would tighten but not qualitatively change these figures.

\subsection*{Mechanism contrast: Experiment 1 vs Experiment 6}
\begin{table}[H]
\centering
\caption*{Table A6. Mechanism contrast, BreastMNIST, 5 seeds, paired: compliance-weighted allocation (Experiment 1) vs uniform server-side DP at the same mean noise scale (Experiment 6). Macro-accuracy \%, mean $\pm$ SD; p values two-sided.}
\small
\begin{tabular}{lccccc}
\toprule
Strategy & E1 mean $\pm$ SD & E6 mean $\pm$ SD & Paired diff (pp) & t & p \\
\midrule
FedAvg & 61.54 $\pm$ 8.38 & 61.72 $\pm$ 8.43 & $-0.18$ & $-0.30$ & 0.782 \\
FedMedian & 64.22 $\pm$ 7.45 & 63.51 $\pm$ 6.93 & +0.71 & 0.50 & 0.646 \\
FedYogi & 51.62 $\pm$ 3.61 & 50.00 $\pm$ 0.00 & +1.62 & 1.00 & 0.374 \\
FedProx & 62.02 $\pm$ 7.93 & 63.68 $\pm$ 7.94 & $-1.67$ & $-1.02$ & 0.365 \\
FedAdam & 50.00 $\pm$ 0.00 & 50.06 $\pm$ 0.14 & $-0.06$ & $-1.00$ & 0.374 \\
\bottomrule
\end{tabular}
\end{table}

\textit{Pooled analysis.} Mean pooled paired difference: +0.09 pp. Sign test: 6 positive / 7 negative / 12 zero, p = 1.000.

\textit{Interpretation.} Compliance-weighted allocation is statistically indistinguishable from uniform server-side DP at the same mean noise scale: no per-strategy paired difference approaches significance, and the pooled difference is +0.09 pp. This is the allocation-neutrality result that grounds the revised framing of the mechanism: compliance weighting neither improves nor degrades utility relative to uniform noise on this benchmark --- its role is governance, giving investigators an auditable, compliance-justified control over per-site perturbation at no measurable utility cost.

\subsection*{Per-client $\sigma_i$ and per-step contribution to the aggregator-dataset cumulative privacy budget}
The $(\varepsilon,\delta)$ bound provided by the framework's DP-SGD step is defined on the aggregator dataset, and under heterogeneous-$\sigma$ composition it is a single cumulative bound across all rounds and per-client DP-SGD applications --- not 16 client-specific bounds. Per-client $\sigma_i$ values calibrate the per-step R\'enyi-DP contribution [4] that each client's update injects into this cumulative aggregator-dataset bound. The table below shows the per-client $\sigma_i$ mapping for representative compliance scores under Equation 2 ($\sigma_{\min} = 0.4$, $\sigma_{\max} = 1.0$), along with the qualitative direction of each client's per-step contribution to the cumulative aggregator-dataset $\varepsilon$. The composed cumulative bounds for the experimental configurations are those of the summary table above ($\varepsilon \approx 1434$ BreastMNIST, $\varepsilon \approx 513$ PneumoniaMNIST at $\delta = 10^{-5}$); configurations dominated by low-compliance (high-$\sigma_i$) clients consume less of the cumulative budget per step, yielding tighter composed bounds.

\begin{table}[H]
\centering
\caption*{Per-client $\sigma_i$ mapping and per-step contribution to the aggregator-dataset cumulative privacy budget.}
\footnotesize
\begin{tabular}{p{0.16\linewidth}p{0.10\linewidth}p{0.13\linewidth}p{0.48\linewidth}}
\toprule
Compliance score ($S_c$) & $\sigma_i$ (from Eq. 2) & Per-step DP-SGD noise multiplier & Per-step contribution to the aggregator-dataset cumulative $\varepsilon$ (qualitative) \\
\midrule
1.00 (fully compliant) & 0.40 & 0.40 & Largest per-step R\'enyi-DP contribution (least noise per step) --- drives the cumulative $\varepsilon$ higher within the composed bound. \\
0.75 & 0.55 & 0.55 & Above-mean per-step contribution; configurations dominated by high-compliance clients yield a looser overall bound. \\
0.50 & 0.70 & 0.70 & Mid-range per-step contribution; balances the composed cumulative bound. \\
0.25 & 0.85 & 0.85 & Below-mean per-step contribution; lower-compliance clients receive more per-step noise and consume less of the cumulative budget per step. \\
0.00 (minimal compliance) & 1.00 & 1.00 & Smallest per-step contribution (most noise per step); configurations dominated by low-compliance clients yield a tighter cumulative bound. \\
\bottomrule
\end{tabular}
\end{table}

Client-data formal DP would require composing the framework with secure aggregation and central client-level Gaussian noise on the aggregated update, establishing $(\varepsilon,\delta)$ at the level of individual client records; this is the principled DP upgrade pathway and remains future work, as does tighter per-step accounting under the Privacy Random Variable (PRV) accountant for heterogeneous-$\sigma$ composition.

\subsection*{Adversarial misreporting: bounded-harm worked example}
The framework's reliance on honest self-reported compliance scores (see Limitations) is partially self-limiting under the aggregator-level $(\varepsilon,\delta)$ bound. The following worked example illustrates the bounded consequence of an over-reporting attack. Suppose institution A's true compliance posture corresponds to $S_c = 0.3$ (substantially below the median), but A reports $S_c = 1.0$ to obtain a lower assigned noise scale and a more favorable contribution weight.

\begin{table}[H]
\centering
\caption*{Bounded-harm worked example: honest reporting vs over-reporting.}
\footnotesize
\begin{tabular}{p{0.24\linewidth}p{0.34\linewidth}p{0.34\linewidth}}
\toprule
Quantity & Under honest reporting (true $S_c = 0.3$) & Under over-reporting (claims $S_c = 1.0$) \\
\midrule
Assigned $\sigma_i$ (from Eq. 2) & 0.82 & 0.40 \\
Per-round noise on A's contribution & Substantial (close to $\sigma_{\max}$) & Minimal ($\sigma_{\min}$) \\
Informal practical protection of A's data via noise injection & Stronger (more noise on A's update) & Weaker (less noise on A's update) \\
A's per-step contribution to the aggregator-dataset cumulative $\varepsilon$ & Smaller per-step share of cumulative budget & Larger per-step share --- accelerates the cumulative aggregator-dataset $\varepsilon$ \\
Net effect of misreporting & --- & Weaker informal protection for A's own data + larger per-step contribution that loosens the shared formal aggregator-dataset bound for the whole federation \\
\bottomrule
\end{tabular}
\end{table}

Three consequences are bounded by the framework's design. (1) The over-reporter receives less noise on its own contribution, reducing the informal practical protection that the framework's noise injection provides for that client's data --- the framework does not provide formal client-level DP, so this informal layer is the only mechanism-level protection client data receive. (2) The misreporter's per-step R\'enyi-DP contribution to the aggregator-dataset cumulative $\varepsilon$ increases (smaller $\sigma \rightarrow$ larger per-step contribution), accelerating consumption of the shared cumulative budget; the result is a looser formal bound on the aggregator dataset for every participant in the federation. (3) The global model incorporates A's contribution with lighter noise, which preserves more signal in A's local data --- but if the over-reporting accompanies the genuinely lower data quality that the true $S_c = 0.3$ would imply, global model accuracy is degraded by A's noisier or biased contribution. The framework does not detect the discrepancy intrinsically. Verification mechanisms identified in Future Work --- on-site automated compliance scanners, cryptographic attestation via Trusted Execution Environment quotes or signed compliance manifests, periodic third-party audits, and game-theoretic incentive design --- are the architecturally appropriate mitigation. The design ensures that the cost of dishonest reporting is distributed across the misreporter (weaker informal protection for its own data) and the federation as a whole (faster consumption of the shared cumulative bound), rather than concentrating the benefit on the misreporter alone.

\subsection*{Reporting conventions and reproducibility}
All per-strategy statistics are paired on matched seeds (seeds 0, 1, 2, 42, and 100 for the five-seed analyses; seeds 0--2 for the three-seed sensitivity analyses). Confidence intervals in Table A2 are t-based with 4 degrees of freedom; all p values are two-sided; zero paired differences are excluded from the Wilcoxon statistic and counted separately in the sign tests, per standard practice [2,3]. Privacy budgets are computed with the Opacus RDP accountant [5] under full sequential composition and reported at $\delta = 10^{-5}$. Every number in this appendix regenerates from the per-seed JSON outputs included in the reviewer repository; this document is built by the repository script scripts/build\_r3\_appendix.py.

\subsection*{Appendix References}
\begin{enumerate}[label={\arabic*.}, leftmargin=*]
\item Yang J, Shi R, Wei D, Liu Z, Zhao L, Ke B, et al. MedMNIST v2 --- a large-scale lightweight benchmark for 2D and 3D biomedical image classification. Sci Data. 2023;10(1):41. doi:10.1038/s41597-022-01721-8
\item Wilcoxon F. Individual comparisons by ranking methods. Biometrics Bulletin. 1945;1(6):80-83. doi:10.2307/3001968
\item Dixon WJ, Mood AM. The statistical sign test. J Am Stat Assoc. 1946;41(236):557-566. doi:10.1080/01621459.1946.10501898
\item Mironov I. R\'enyi differential privacy. In: Proceedings of the IEEE 30th Computer Security Foundations Symposium (CSF); 2017:263-275. doi:10.1109/CSF.2017.11
\item Yousefpour A, Shilov I, Sablayrolles A, Testuggine D, Prasad K, Malek M, et al. Opacus: user-friendly differential privacy library in PyTorch. arXiv:2109.12298 [preprint]; 2021.
\end{enumerate}

\end{document}